%% file: main.tex
\documentclass{article}
\usepackage{arxiv}
\usepackage[utf8]{inputenc} 
\usepackage[T1]{fontenc}    
\usepackage{url}            
\usepackage{booktabs}       
\usepackage{amsfonts}       
\usepackage{nicefrac}       
\usepackage{microtype}      
\usepackage{lipsum}
\usepackage{amsmath}
\usepackage{amsthm}
\usepackage{amssymb}   
\usepackage{bigints}
\usepackage{bm}
\usepackage{diagbox}
\usepackage{makecell}
\usepackage{multirow}
\usepackage{caption}
\usepackage{subcaption}
\usepackage{graphicx}
\usepackage[ruled,vlined]{algorithm2e}
\include{pythonlisting}
\usepackage{xcolor}
\usepackage{mathtools}
\usepackage{enumitem}

\usepackage{hyperref}
\hypersetup{
    colorlinks=true,
    citecolor=black,
    linkcolor=black,
    filecolor=black,
    urlcolor=black,
}

\theoremstyle{definition}

\numberwithin{equation}{section}

\theoremstyle{plain}

\newtheorem{theorem}{Theorem}[section]
\newtheorem{definition}[theorem]{Definition}

\title{Respecting causality is all you need for training physics-informed neural networks}

\author{
  Sifan Wang \\
  Graduate Group in Applied Mathematics \\
  and Computational Science \\
  University of Pennsylvania\\
  Philadelphia, PA 19104 \\
  \texttt{sifanw@sas.upenn.edu} \\
   \And
  Shyam Sankaran \\
  Department of Mechanical Engineering \\
  and Applied Mechanics \\
  University of Pennsylvania\\
  Philadelphia, PA 19104 \\
  \texttt{shyamss@seas.upenn.edu} \\
  \And
  Paris Perdikaris \\
  Department of Mechanical Engineering \\
  and Applied Mechanics\\
  University of Pennsylvania\\
  Philadelphia, PA 19104 \\
  \texttt{pgp@seas.upenn.edu} \\
}

\begin{document}
\maketitle

\begin{abstract}
While the popularity of physics-informed neural networks (PINNs) is steadily rising, to this date PINNs have not been successful in simulating dynamical systems whose solution exhibits multi-scale, chaotic or turbulent behavior. In this work we attribute this shortcoming to the inability of existing PINNs formulations to respect the spatio-temporal causal structure that is inherent to the evolution of physical systems. We argue that this is a fundamental limitation and a key source of error that can ultimately steer PINN models to converge towards erroneous solutions. We address this  pathology by proposing a simple re-formulation of PINNs loss functions that can explicitly account for physical causality during model training. We demonstrate that this simple modification alone is enough to introduce significant accuracy improvements, as well as a practical quantitative mechanism for assessing the convergence of a PINNs model. We provide state-of-the-art numerical results across a series of benchmarks for which existing PINNs formulations fail, including the chaotic Lorenz system, the Kuramoto–Sivashinsky equation in the chaotic regime, and the Navier-Stokes equations in the turbulent regime. To the best of our knowledge, this is the first time that PINNs have been successful in simulating such systems, introducing new opportunities for their applicability to problems of industrial complexity.
\end{abstract}


\section{Introduction}

Physics-informed neural networks (PINNs) have emerged as a promising framework for synthesizing observational data and physical laws across diverse applications in science and engineering \cite{raissi2020hidden, mathews2021uncovering, kissas2020machine,yazdani2020systems, wang2021deep, shukla2020physics,chen2020physics, sahli2020physics}. However, it is well known that PINNs often face severe difficulties and even fail to tackle problems whose solution exhibits highly nonlinear, multi-scale, or chaotic behavior \cite{wang2021eigenvector, karniadakis2021physics}.
Over the last few years, a series of extensions to the original formulation of Raissi {\it et al.} \cite{raissi2019physics} have been proposed with the sole goal of enhancing the accuracy and robustness of PINNs in tackling increasingly more challenging problems. Such extensions include, but are not limited to, novel optimization algorithms for adaptive training \cite{wang2021understanding, wang2022and, mcclenny2020self, maddu2021inverse}, adaptive algorithms for selecting batches of training data \cite{wight2020solving, nabian2021efficient}, novel network architectures \cite{wang2021understanding, wang2021eigenvector,bu2021quadratic,jagtap2022deep, liang2021reproducing}, domain decomposition strategies \cite{jagtap2020extended,moseley2021finite}, new types of activation functions \cite{jagtap2020adaptive}, and sequential learning strategies \cite{wight2020solving, krishnapriyan2021characterizing, mattey2022novel}. Although these techniques have been successful in introducing some improvements in terms of trainability and accuracy, there still exists a vast suite of problems that  remain elusive to PINNs. Examples of such problems include systems whose behavior exhibits strong non-linearity, broadband energy spectra, and high sensitivity to initial conditions, such as the chaotic Kuramoto-Sivishinski equation and the Navier-Stokes equations in the turbulent regime. These are not pathological corner cases, but cases that are extremely relevant across a multitude of realistic scenarios in science and engineering. Therefore, there is a pressing need for understanding why PINNs fall short in such scenarios, and how they can be improved in order to overcome the challenges that currently limit their success to relatively simple problems.

Physical systems are known to possess  an inherent causal structure. Consider for example a linear wave with some initial velocity that is spreading out with a speed $c$ across a homogeneous medium \cite{strauss2007partial}. It is well-understood that, although a part of the wave may lag behind (if there is an initial velocity), no part can travel faster than speed $c$. This assertion encapsulates the so-called {\em principle of causality} that dictates how local changes in the initial/boundary data of a spatio-temporal dynamical system is reflected in its corresponding states at later times \cite{strauss2007partial}. Specific to hyperbolic partial differential equations (PDEs), such as the wave equation, this principle underpins the formulation of the method of characteristics \cite{evans1998partial} that provides a rigorous set of analytical and numerical tools for efficiently tackling initial value problems. Although characterizing how information propagates in general nonlinear PDEs is a challenging task, basic principles of causality such as temporal precedence and covariation (i.e. statistical dependency between variables that are generated by coupled time evolution) are  still expected to hold. This causal structure is also clearly reflected in classical numerical methods, where a PDE is typically discretized in time by sequential algorithms which ensure that the solution at time $t$ is fully resolved before approximating the solution at time $t+\Delta t$. Strikingly, this notion of temporal dependence is absent in most continuous-time PINNs formulations (see e.g.  \cite{lagaris1998artificial, raissi2019deep, kharazmi2019variational, jagtap2020extended, wang2021understanding, wang2022and, jagtap2020adaptive}). In fact, as we will see in section \ref{sec:causal_training}, continuous-time PINNs trained by gradient descent are implicitly biased towards first approximating PDE solutions at later times, before even resolving the initial conditions, therefore profoundly violating temporal causality. Consequently, it is no surprise that such formulations are fragile and often fail to simulate forward problems, especially in cases where the target solutions exhibit strong dependence on initial data (e.g. chaotic systems). Recent studies \cite{wight2020solving, krishnapriyan2021characterizing, mattey2022novel} have proposed remedies to this issue by empirically introducing sequential training strategies, yet a concrete justification of why such strategies appear to be effective is still missing.

This work is focused on investigating the importance of respecting physical causality during the training of continuous-time PINNs.
Our specific contributions can be summarized as:
\begin{itemize}
\item We reveal an implicit bias suggesting that continuous-time PINNs models can violate causality, and hence are susceptible to converge towards erroneous solutions.
\item We put forth a simple re-formulation of PINNs loss functions that allows us to explicitly respect the causal structure that characterizes the solution of general nonlinear PDEs.
\item Strikingly, we demonstrate that this simple modification alone is enough to introduce significant accuracy improvements, allowing us to tackle problems that have remained elusive to PINNs.
\item We provide a practical quantitative criterion for assessing the training  convergence of a PINNs model.
\item We examine a collection of challenging benchmarks for which existing PINNs formulations fail, and demonstrate that the proposed {\em causal training} strategy leads to state-of-the-art results.
\end{itemize}
To the best of our knowledge, this is the first time that PINNs have been successful in simulating systems such as the chaotic Lorenz system, the Kuramoto–Sivashinsky equation in the chaotic regime, and the Navier-Stokes equations in the turbulent regime, introducing new opportunities for their applicability to problems of industrial complexity.

The paper is structured as follows. In section \ref{sec: pinns},  we provide an overview of PINNs following the original formulation of Raissi {\em et. al.} \cite{raissi2019physics}. Using a simple case study, we reveal an implicit bias of continuous-time PINNs that makes them prone to violating physical causality, and thereby steering them towards erroneous solutions. To address this drawback, in section \ref{sec:causal_training} we put forth a simple re-formulation of the PINNs residual loss and propose a general {\em casual training} algorithm for explicitly respecting physical causality during model training. Section \ref{sec:practical} discusses practical considerations specific to enhancing the accuracy and efficiency of PINNs. These developments are put to test in section \ref{sec:results}, where we demonstrate state-of-the-art results across a comprehensive collection of challenging benchmarks for which existing PINN formulations are known to fail. Finally, section \ref{sec:discussion} provides a summary of our main findings, and touches upon remaining limitations and areas for future research.

\section{Physics-informed neural networks (PINNs)}
\label{sec: pinns}

\paragraph{Problem setup:} 
We begin with a brief overview of physics-informed neural networks (PINNs) \cite{raissi2019physics} in the context of inferring the solutions of PDEs. Generally, we consider PDEs taking the form
\begin{align}
\label{eq: PDE}
     \bm{u}_t +  \mathcal{N}[\bm{u}] = 0, \ \  t \in [0, T],  \ \bm{x} \in \Omega,
\end{align} 
subject to the initial and boundary conditions
\begin{align}
     \label{eq: IC}
     &\bm{u}( 0, \bm{x})=\bm{g}(\bm{x}), \ \ \bm{x} \in \Omega, \\
      \label{eq: BC}
     &\mathcal{B}[\bm{u}] = 0,  \ \   t\in [0, T], \  \bm{x} \in  \partial \Omega,
\end{align}
where $\mathcal{N}[\cdot]$ is a linear or nonlinear differential operator, and  $\mathcal{B}[\cdot] $ is a boundary operator  corresponding to Dirichlet, Neumann, Robin, or periodic boundary conditions. In addition, $\bm{u}$ describes the unknown latent solution that is governed by the  PDE system of Equation \ref{eq: PDE}. 

Following the original work of Raissi {\em et al.} \cite{raissi2019physics}, we proceed by representing the unknown solution $\bm{u}(t, \bm{x})$ by a deep neural network $\bm{u}_{\bm{\theta}}(t, \bm{x})$, where $\bm{\theta}$ denotes all tunable parameters of the network (e.g., weights and biases). Then, a physics-informed model can be trained by minimizing the following composite loss function
\begin{align}
    \label{eq: PINN_loss}
    \mathcal{L}(\bm{\theta}) =  \lambda_{ic} \mathcal{L}_{ic}(\bm{\theta}) + \lambda_{bc} \mathcal{L}_{bc}(\bm{\theta}) +  \lambda_r \mathcal{L}_r(\bm{\theta}), 
\end{align}
where 
\begin{align}
    \label{eq: loss_ic}
     &\mathcal{L}_{ic}(\bm{\theta}) = \frac{1}{N_{ic}} \sum_{i=1}^{N_{ic}} \left| \bm{u}_{\bm{\theta}}(0, \bm{x}_{ic}^i) - \bm{g}(\bm{x}_{ic}^i) \right|^2, \\
     \label{eq: loss_bc}
     &\mathcal{L}_{bc}(\bm{\theta}) = \frac{1}{N_{bc}} \sum_{i=1}^{N_{bc}} \left| \mathcal{B}[\bm{u}_{\bm{\theta}}]( t_{bc}^i, \bm{x}_{bc}^i) \right|^2, \\
    \label{eq: loss_r}
    &\mathcal{L}_r(\bm{\theta}) = \frac{1}{N_r} \sum_{i=1}^{N_r} \left| \frac{\partial \bm{u}_{\bm{\theta}}}{\partial t}(t_r^i, \bm{x}_r^i) + \mathcal{N}[\bm{u}_{\bm{\theta}}](t_r^i, \bm{x}_r^i) \right|^2. 
\end{align}
Here $\{\bm{x}_{ic}^i\}_{i=1}^{N_{ic}}$, $\{t_{bc}^i, \bm{x}_{bc}^i\}_{i=1}^{N_{bc}}$ and $\{t_{r}^i, \bm{x}_{r}^i\}_{i=1}^{N_{r}}$ can be the vertices of a fixed mesh or points that are randomly sampled at each iteration of a gradient descent algorithm. Notice that all required gradients with respect to input variables or network parameters $\bm{\theta}$ can be efficiently computed via automatic differentiation \cite{griewank2008evaluating}. Moreover, the hyper-parameters $\left\{  \lambda_{ic},  \lambda_{bc}, \lambda_r \right\}$ allow the flexibility of assigning a different learning rate to each individual loss term in order to balance their interplay during model training. These weights may be user-specified or tuned automatically during training \cite{wang2021understanding, wang2022and}. 

\paragraph{An illustrative example:} 
To motivate the proposed methods described in section \ref{sec:causal_training}, let us study a representative case with which conventional PINN models are known to struggle. To this end, consider the one-dimensional Allen-Cahn equation 
\begin{align}
    &u_{t}-0.0001 u_{x x}+5 u^{3}-5 u=0, \quad t \in[0,1], x \in[-1,1], \\
    &u(x, 0)=x^{2} \cos (\pi x), \\
    &u(t, -1)=u(t, 1), \\
    &u_{x}(t, -1)=u_{x}(t, 1).
\end{align}
This example is difficult to directly solve with the original continuous-time formulation of Raissi {et al.} \cite{raissi2019physics}, and has been recently studied by {\em Wight et. al.} \cite{wight2020solving} and {\em McClenny et. al.} \cite{mcclenny2020self} who developed adaptive re-sampling and weighting algorithms, respectively, to improve the PINNs prediction. 

Following the setup discussed in these studies \cite{mcclenny2020self, wight2020solving}, we 
represent the latent variable $u$ by a fully-connected neural network $u_{\bm{\theta}}$ with tanh activation function, 4 hidden layers and 128 neurons per hidden layer. To further simplify the training objective \ref{eq: PINN_loss}, we also strictly impose the periodic BCs by embedding the input coordinates into Fourier expansion using Equation \ref{eq: 1D_Fourier} with $m = 10$ (see section \ref{sec:practical} for further details). Then the loss function \ref{eq: PINN_loss} can be reduced to
 \begin{align}
        \mathcal{L}(\bm{\theta}) =  \lambda_{ic} \mathcal{L}_{ic}(\bm{\theta})  +  \lambda_r \mathcal{L}_r(\bm{\theta}), 
\end{align}
where  $\mathcal{L}_{ic}(\bm{\theta})$ and $\mathcal{L}_r(\bm{\theta}) $ are defined exactly the same as in Equation \ref{eq: loss_ic} and Equation \ref{eq: loss_r}. For simplicity, we create a uniform mesh of size $100 \times 256$  in the computational domain $[0, 1] \times [-1, 1]$, yielding $N_{ic} = 256$ initial points  and $N_r = 25600$ collocation points for enforcing the PDE residual. We also choose $\lambda_{ic} = 100, \lambda_{r} = 1$ for better enforcing the initial condition.

\begin{figure}
     \centering
     \begin{subfigure}[b]{0.8\textwidth}
         \centering
         \includegraphics[width=\textwidth]{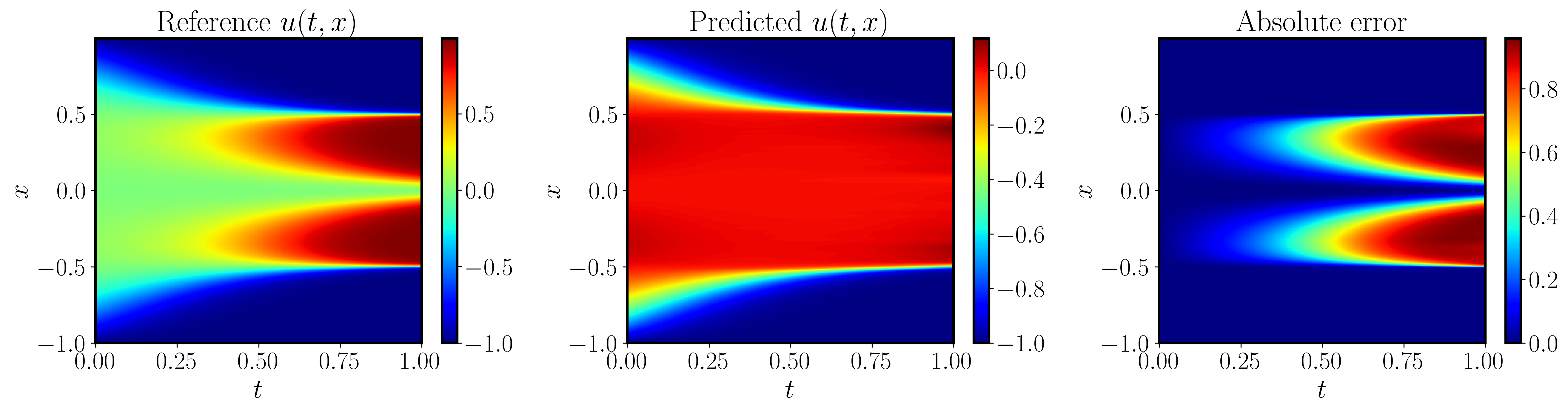}
     \end{subfigure}
     \begin{subfigure}[b]{0.7\textwidth}
         \centering
         \includegraphics[width=\textwidth]{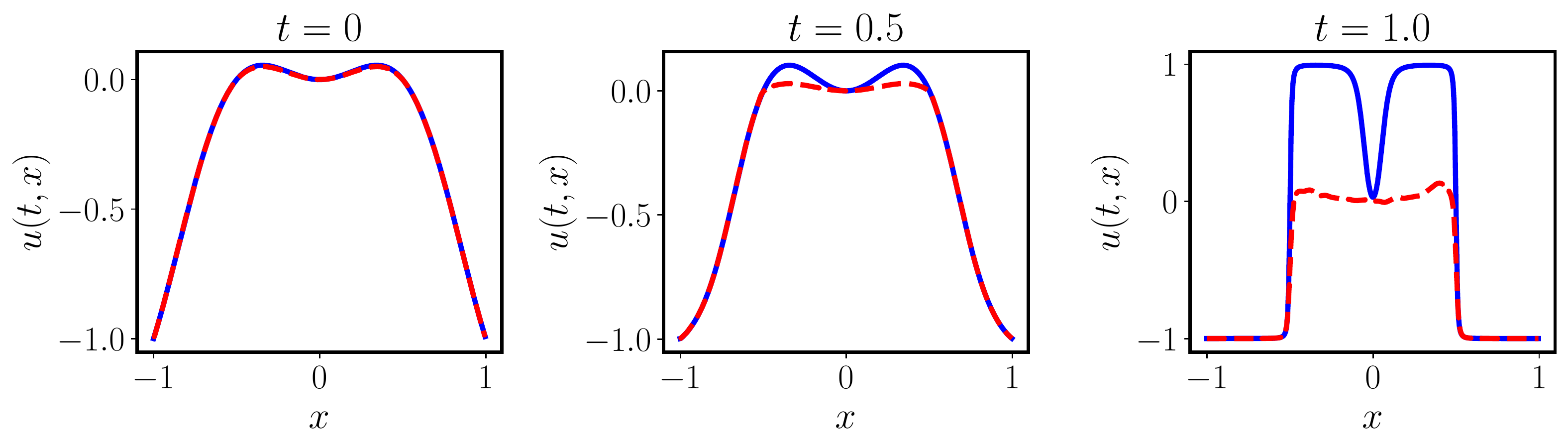}
     \end{subfigure}
        \caption{{\em Allen-Cahn equation:} {\em Top:}  Reference solution versus the prediction of a trained conventional physics-informed neural network. The resulting relative $L^2$ error is $49.87\%$.  {\em Bottom:}  Comparison of the predicted and reference solutions corresponding to the three temporal snapshots at $t=0.0, 0.5, 1.0$. }
        \label{fig: AC_vanilla_PINN_pred}
\end{figure}

\begin{figure}
    \centering
    \includegraphics[width=0.9\textwidth]{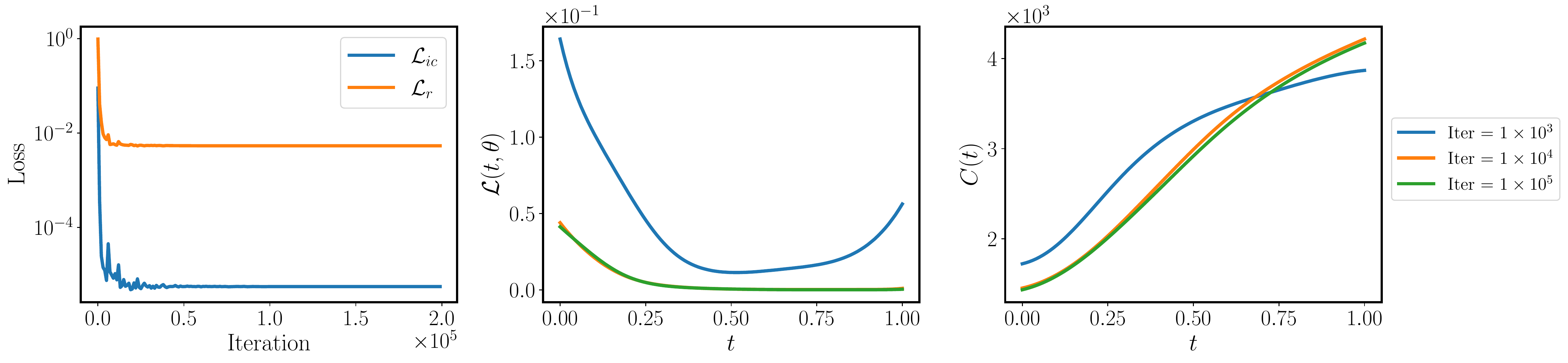}
       \caption{{\em Allen-Cahn equation:} {\em Left:} Loss convergence of training a conventional physics-informed neural network for $2 \times 10^5$ iterations.  {\em Middle:} Temporal residual loss $\mathcal{L}(t, \bm{\theta})$ at different iteration of the training. {\em Right:} Temporal convergent rate at different iteration of the training.}
    \label{fig: AC_vanilla_PINN_loss}
\end{figure}

We proceed by training the resulting PINN model via full-batch gradient descent using the Adam optimizer \cite{kingma2014adam} for $2 \times 10^5$ iterations. As shown in  Figure \ref{fig: AC_vanilla_PINN_pred}, even when the periodic boundary conditions are enforced exactly, our conventional PINN model is unable to learn the accurate solution for this example. One can also observe that the predicted solution seems to get stuck at some intermediate state and cannot be further refined to provide an accurate approximation to the ground truth. This is consistent with the left panel of Figure \ref{fig: AC_vanilla_PINN_loss} where the loss functions rapidly decrease in the first few thousand training iterations, and then barely change for the rest of training, implying that the neural network gets trapped in an erroneous local minimum. Unfortunately, such problematic behavior is not a rare event, but rather a common outcome for PINNs, especially when solving transient problems \cite{wang2022and,krishnapriyan2021characterizing}. 

\paragraph{PINNs can violate physical causality:}
To explore the underlying reasons behind this failed case study, let us closely examine the definition of the residual loss $\mathcal{L}_r$. Before doing so, we will slightly change our notation for convenience. Suppose that  $0 = t_1 < t_2  < \cdots < t_{N_t} = T$ discretizes the temporal domain, and $\{\bm{x}_j\}_{j=1}^{N_x}$ discretizes the spatial domain $\Omega$. For this example, $\{ t_i\}_{i=1}^{N_t}$ and $\{\bm{x}_j\}_{j=1}^{N_x}$ are uniformly spaced meshes in $[0, 1]$ and $[-1, 1]$, respectively.  Now for a given spatial discretization  $\{\bm{x}_j\}_{j=1}^{N_x}$, we define the temporal residual loss as 
\begin{align}
    \label{eq: temporal_residual}
    \mathcal{L}_r(t, \bm{\theta}) = \frac{1}{N_x} \sum_{j=1}^{N_x} | \frac{\partial \bm{u}_{\bm{\theta}}}{\partial t}(t, \bm{x}_j) + \mathcal{N}[\bm{u}_{\bm{\theta}}](t, \bm{x}_j) |^2. 
\end{align}
Then, the residual loss \ref{eq: loss_r} can be rewritten as 
\begin{align}
    \mathcal{L}_r(\bm{\theta}) &=   \frac{1}{N_t} \sum_{i=1}^{N_t} \mathcal{L}_r(t_i, \bm{\theta})  \\
    &=  \frac{1}{N_t N_x} \sum_{i=1}^{N_t} \sum_{j=1}^{N_x} | \frac{\partial \bm{u}_{\bm{\theta}}}{\partial t}(t_i, \bm{x}_j) + \mathcal{N}[\bm{u}_{\bm{\theta}}](t_i, \bm{x}_j) |^2. 
\end{align}
Next, we discretize $\frac{\partial \bm{u}_{\bm{\theta}}}{\partial t}$ using the forward Euler scheme \cite{iserles2009first}. For any $1 \leq i \leq N_t -1$, $\mathcal{L}(t_i, \bm{\theta}) $  can be approximated by
\begin{align}
    \mathcal{L}_r(t_i, \bm{\theta})  &\approx  \frac{1}{N_x} \sum_{j=1}^{N_x} \left|  \frac{\bm{u}_{\bm{\theta}}(t_{i}, \bm{x}_j) - \bm{u}_{\bm{\theta}}(t_{i-1}, \bm{x}_j)}{\Delta t}  + \mathcal{N}[\bm{u}_{\bm{\theta}}](t_{i}, \bm{x}_j) \right
    |^2 \nonumber \\
    &\approx \frac{|\Omega|}{\Delta t^2} \int_\Omega |\bm{u}_{\bm{\theta}}(t_{i}, \bm{x}) - \bm{u}_{\bm{\theta}}(t_{i-1}, \bm{x}) + \Delta t \mathcal{N}[\bm{u}_{\bm{\theta}}](t_{i}, \bm{x})   |^2 d\bm{x}.
\end{align}
From the above expression, we immediately obtain that the minimization of $\mathcal{L}(t_i, \bm{\theta})$ should be based on the correct prediction of both  $\bm{u}_{\bm{\theta}}(t_{i}, \bm{x})$ and $\bm{u}_{\bm{\theta}}(t_{i-1}, \bm{x})$,  while the original formulation of Equation \ref{eq: loss_r}
tends to minimize all $\mathcal{L}(t_i, \bm{\theta})$ simultaneously. As a result, by using Equation \ref{eq: loss_r}, the residual loss $\mathcal{L}_r(t_i, \bm{\theta})$ will be minimized even if the predictions at $t_i$ and previous times are inaccurate. This behavior inevitably violates temporal causality, making the PINN model susceptible to learn erroneous solutions.

This conclusion is further confirmed by the middle panel of Figure \ref{fig: AC_vanilla_PINN_loss} where we plot the temporal residual loss of Allen-Cahn equation at different iterations of training. As expected, the residual is quite large near the initial state and rapidly decays to nearly zero after $t=0.5$. We emphasize that the PDE temporal residual of small magnitude is meaningful only if the PINN model is well optimized and able to yield accurate predictions at the previous time steps.

\paragraph{An undesirable implicit bias:}
To provide a deeper understanding of the fact that PINNs may violate temporal causality, we analyze their training dynamics through the lens of their empirical Neural Tangent Kernel (NTK) \cite{jacot2018neural, wang2022and}. 
Specifically, for every $\mathcal{L}_r(t, \bm{\theta})$ (Equation \ref{eq: temporal_residual}),  we can define the empirical NTK $\bm{K}_{\bm{\theta}}(t) \in \mathbb{R}^{N_x \times N_x}$ whose $ij$-th entry is given by  \cite{wang2022and}
\begin{align}
    \bm{K}_{\bm{\theta}}(t)_{ij} = \left\langle \frac{\partial \mathcal{R}_{\bm{\theta}} }{\partial \bm{\theta}}(t, \bm{x}_i)  , \frac{\partial \mathcal{R}_{\bm{\theta}} }{\partial \bm{\theta}} (t, \bm{x}_j) \right\rangle,
\end{align}
where $\mathcal{R}_{\bm{\theta}}$ is the corresponding PDE residual defined by
\begin{align}
    \mathcal{R}_{\bm{\theta}}(t, \bm{x}) =   \frac{\partial \bm{u}_{\bm{\theta}}}{\partial t}(t, \bm{x}) + \mathcal{N}[\bm{u}_{\bm{\theta}}](t, \bm{x}), \quad i,j=1,2, \dots, N_x.
\end{align}
As demonstrated by Wang {\em et. al. } \cite{wang2022and}, the eigenvalues of $\bm{K}_{\bm{\theta}}(t)$ determine the convergence rate of each $\mathcal{L}_r(t, \bm{\theta})$ contributing to the total residual loss $\mathcal{L}_r(\bm{\theta})$. Specifically, larger eigenvalues implies faster convergence rate. Following \cite{wang2022and}, we introduce the definition 
\begin{definition}
For any given $t \in [0, T]$,  the temporal convergence rate $C(t)$  of $\mathcal{L}_r(t, \bm{\theta})$ is defined by
\begin{align}
\label{eq:convergence_rate}
    C(t) = \frac{\sum_{k=1}^{N_t} \lambda_k(t) }{N_t} = \frac{\text{Trace}(\bm{K}_{\bm{\theta}}(t))}{N_t},
\end{align}
where $\{\lambda_k(t)\}_{k=1}^{N_t}$ are the eigenvalues of $\bm{K}_{\bm{\theta}}(t)$.
\end{definition}

Equipped with definition \ref{eq:convergence_rate}, we visualize $C(t)$ at different iterations during the training of our PINNs model for solving Allen-Cahn equation. In the right panel of Figure \ref{fig: AC_vanilla_PINN_loss}, it can be seen that $C(t)$ is greater if $t$ is greater, indicating that the network is biased towards minimizing the temporal residual $\mathcal{L}_r(t, \bm{\theta})$ for larger $t$. This reveals an undesirable implicit bias of continuous-time PINN models trained via gradient descent, suggesting that such models can profoundly violate the temporal causal structure that is inherent to time-dependent PDE systems. We argue that this inherent pathology of PINNs is the key underlying reason behind their inability to simulate transient problems that exhibit strong temporal correlations and sensitivity to initial data.

In the next section we put forth a remarkably simple and effective strategy for explicitly respecting physical causality during the training phase PINNs.


\section{Causal training for physics-informed neural networks}
\label{sec:causal_training}

\paragraph{A simple re-formulation:} 
Based on our findings in the previous section, it is natural to ask how we can respect physical causality when solving PDEs with PINNs.
We answer this question by introducing a simple re-formulation of the PINNs training objective that can explicitly account for the missing causal structure. To this end, we define a weighted residual loss as
\begin{align}
    \label{eq: weighted_residual_loss}
         \mathcal{L}_r(\bm{\theta}) =   \frac{1}{N_t} \sum_{i=1}^{N_t}  w_i \mathcal{L}_r(t_i, \bm{\theta}).
\end{align}
We recognize that the weights $w_i$ should be large -- and therefore allow the minimization of $\mathcal{L}_r(t_i, \bm{\theta})$ -- only if all residuals $\{\mathcal{L}_r(t_k, \bm\theta)\}_{k=1}^i$ before $t_i$ are minimized properly,  and vice versa. This can be achieved by expressing the weights $w_i$ as
\begin{align}
    \label{eq: temporal_weights}
     w_i = \exp\left(- \epsilon \sum_{k=1}^{i-1}  \mathcal{L}_r(t_k, \bm{\theta})\right)    , \text{ for } i = 2, 3, \dots, N_t,
\end{align}
where $\epsilon$ will be referred to as a {\em causality parameter} that controls the steepness of the weights $w_i$ (see below for a more detailed discussion).
As such, the weighted residual loss can be written as
\begin{align}
    \label{eq: weighted_residual}
    \mathcal{L}_r(\bm{\theta}) =   \frac{1}{N_t} \sum_{i=1}^{N_t}\exp\left(- \epsilon \sum_{k=1}^{i-1}  \mathcal{L}_r(t_k, \bm{\theta})\right)    \mathcal{L}_r(t_i, \bm{\theta}).
\end{align}
Notice that $w_i$ is inversely exponentially proportional to the magnitude of the cumulative residual loss from the previous time steps. As a consequence, $\mathcal{L}_r(t_i, \bm{\theta})$ will not be minimized unless all previous residuals $\{\mathcal{L}_r(t_k, \bm{\theta})\}_{k=1}^{i-1}$ decrease to some small value such that $w_i$ is large enough.

\begin{figure}
     \centering
     \begin{subfigure}[b]{0.8\textwidth}
         \centering
         \includegraphics[width=\textwidth]{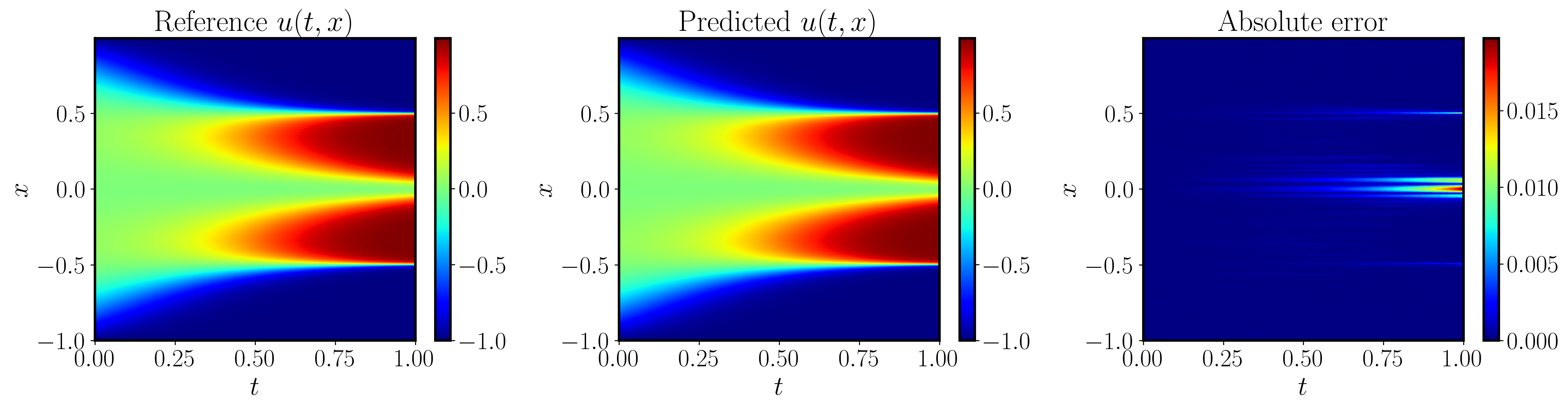}
     \end{subfigure}
     \begin{subfigure}[b]{0.7\textwidth}
         \centering
         \includegraphics[width=\textwidth]{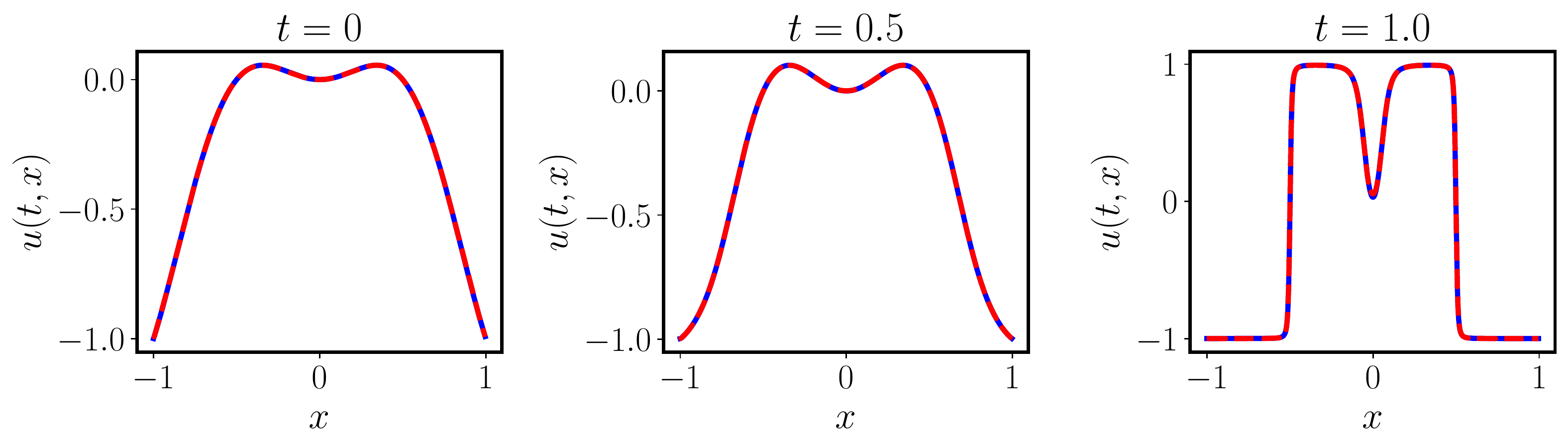}
     \end{subfigure}
        \caption{{\em Allen-Cahn equation:} {\em Top:}  Reference solution versus the prediction of a trained physics-informed neural network using  Algorithm \ref{alg}. The resulting relative $L^2$ error is $1.43e-03$.  {\em Bottom:}  Comparison of the predicted and reference solutions corresponding to the three temporal snapshots at $t=0.0, 0.5, 1.0$. }
        \label{fig: AC_TW_PINN_pred}
\end{figure}

\begin{figure}
    \centering
    \includegraphics[width=0.9\textwidth]{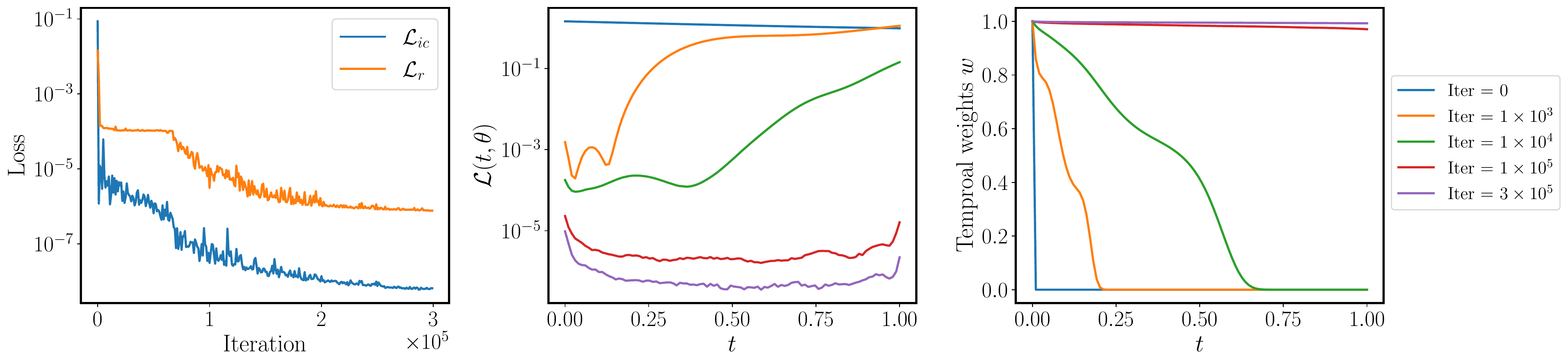}
       \caption{{\em Allen-Cahn equation:} {\em Left:} Loss convergence of training a physics-informed neural network  using  Algorithm \ref{alg}.  {\em Middle:} Temporal residual loss $\mathcal{L}(t, \bm{\theta})$ at different iteration of the training. {\em Right:} Temporal weights at different iteration of the training.}
    \label{fig: AC_TW_PINN_loss_weights}
\end{figure}

We now employ this simple modification and revisit the Allen-Cahn case study discussed before. We proceed by training the same network by minimizing the loss of Equation \ref{eq: PINN_loss} using the weighted residual loss of Equation \ref{eq: weighted_residual} with $\epsilon = 100$,  for $3 \times 10^5$ iterations of gradient descent under exactly the same hyper-parameter settings.  The results of this experiment are summarized in Figure \ref{fig: AC_TW_PINN_pred}. One can see that the predicted solution achieves an excellent agreement with the ground truth, yielding an approximation error of $1.43e-03$ measured in the relative $L^2$ norm. 
The left panel of Figure \ref{fig: AC_TW_PINN_loss_weights} presents the convergence of the different loss function components, which is evidently  much better than the one presented in Figure \ref{fig: AC_vanilla_PINN_loss}. Here we note that no other modifications between the two cases exist, besides the use of the proposed weighted residual loss of Equation \ref{eq: weighted_residual}. 

In fact, if in conjunction with the weighted residual loss we also employ a more powerful architecture for this example, such as the modified MLP  \cite{wang2021understanding} described in section \ref{sec: modified_mlp}, then we can achieve an even more accurate result with a resulting relative $L^2$ error of $1.39e-04$. Additional detailed visual assessments for this example are provided in Appendix \ref{appendix: AC}. 

Finally, in Table \ref{tab: AC} we provide the accuracy reported for this problem by existing approaches in the literature  \cite{mcclenny2020self, wight2020solving, mattey2022novel}. It is evident that the proposed methodology outperforms the best reported result of competing approaches by a factor of $\sim$10-100x. This is a strong indication of the significance and necessity of respecting causality in training PINNs. 

\begin{table}
    \renewcommand{\arraystretch}{1.4}
    \centering
    \begin{tabular}{l|c}
    \hline
    Method   & Relative $L^2$ error  \\
     \hline
      Original formulation of Raissi {\it et al.} \cite{raissi2019physics}    &  $4.98e-01$ \\
      Adaptive time sampling \cite{wight2020solving} & $2.33e-02$ \\
       Self-attention \cite{mcclenny2020self} & $2.10e-02$  \\
       Time marching \cite{mattey2022novel}  & $1.68e-02$ \\
       {\bf Causal training (MLP)} & $\mathbf{1.43e-03}$ \\
     {\bf Causal training (modified MLP)}  & $\mathbf{1.39e-04}$ \\
    \hline
    \end{tabular}
    \caption{{\em Allen-Cahn equation:} Relative $L^2$ errors obtained by different approaches.}
    \label{tab: AC}
\end{table}

\paragraph{A stopping criterion for assessing training convergence:}
To understand the effect of the residual weights $\{w_i\}$, we present the temporal residual loss and weights at different iterations of gradient descent in the middle and right panel of Figure \ref{fig: AC_TW_PINN_loss_weights} and Figure \ref{fig: AC_TW_modified_MLP_PINN_loss_L_t_weights}. We observe that the initial temporal weights are all zero except for $t=0$, implying that only $\mathcal{L}_r(t_0, \bm{\theta})$ will be minimized at the beginning of training. Throughout the rest of the training, more temporal weights are activated, and eventually, all of them converge to $1$ as the PDE residual loss is properly minimized. This last observation suggests that monitoring the magnitude of the residual weights $\{w_i\}$ can provide an effective 
 stopping criterion for assessing the convergence of a PINNs model during training. Specifically, one can choose to terminate training of $\min_i w_i > \delta$, for some chosen threshold parameter $\delta \in (0,1)$. 
 As we will see in section \ref{sec:results}, this stopping criterion not only helps to train a PINNs model faster, but it actually yields trained models with superior predictive accuracy.
 
\paragraph{Sensitivity on the causality parameter $\epsilon$:}
Here we must note that the results obtained using the proposed weighted residual loss do exhibit some sensitivity to the causality parameter $\epsilon$ in Equation \ref{eq: temporal_weights}. Choosing a very small $\epsilon$  can prevent the network from effectively minimizing the latter temporal residuals. On the other hand, choosing a large $\epsilon$ value can result in a more difficult optimization problem, because the temporal residuals at earlier times have to decrease to a very small value in order to activate the latter temporal weights. This may be hard to achieve in some cases due to limited network capacity in minimizing the target residuals.
In order to avoid tedious hyper-parameter tuning, we employ an annealing strategy for adjusting $\epsilon$ using an increasing sequence of values $\{\epsilon_i\}_{i=1}^k$, which gradually increases the strength with which the PDE residual constraint is enforced. As we will see in section \ref{sec:results}, we empirically observe that this choice yields the best results in practice.

\paragraph{Fitting the initial data:}\label{par:ics}
In the spirit of respecting causality, one may recognize that all temporal residuals should be minimized only if the network can first accurately fit the initial data. Therefore, we may treat the initial loss $\mathcal{L}_{ic}$ as a special temporal residual at $t=0$ and incorporate it into the weighted residual loss of Equation \ref{eq: weighted_residual_loss} in the same manner. 

\paragraph{Causal training for PINNs:}
Based on the above remarks, Algorithm \ref{alg} presents a general {\it causal training} algorithm for PINNs. Specifically, it summarizes the proposed re-formulation of the residual and initial conditions loss, the annealing scheme for the $\epsilon$ parameter, and the stopping criterion for terminating the training upon the convergence of the temporal weights  $w_i$. 
Accompanying Algorithm \ref{alg}, here we present a few additional remarks worth discussing.
\begin{enumerate}[leftmargin=*]
    \item Although in this work we have limited our attention to PDEs with periodic boundary conditions that can be enforced in an exact manner (see section \ref{sec:practical} for more details), the proposed {\em causal training} algorithm can be adapted to also incorporate boundary constraints using a similar treatment to the initial conditions loss.
    \item Note that the temporal weights $\{w_i\}_{i=0}^{N_t}$ are a function of the trainable parameters $\bm{\theta}$. We use \texttt{lax.stop\_gradient} in our JAX \cite{jax2018github} implementation to prevent gradient back-propagation through the computation of $w_i$.
    \item  The computational cost of the proposed algorithm is negligible compared to conventional PINNs formulations since the weights $w_i$ are computed by directly evaluating the PINNs loss functions, whose values are already stored in the computational graph during training.   
    \item  The proposed algorithm is not limited to fixed mesh points for evaluating the PINNs loss terms, and the collocation points can be randomly sampled at each iteration of gradient descent. The only requirement is that the sampled temporal points $\{t_i\}_{i=1}^{N_t}$ should form a non-decreasing sequence in temporal domain so that temporal causality can be respected.
\end{enumerate}

Here we should also mention that Algorithm \ref{alg} is general and can be employed within any existing physics-informed machine learning pipeline, including physics-informed neural networks \cite{raissi2019physics, lu2019deepxde, kharazmi2019variational, jagtap2022deep, jagtap2020extended, hennigh2021nvidia}, physics-informed deep operator networks \cite{wang2021learning, wang2021long, wang2021improved}, and physics-informed neural operators \cite{li2021physics}.

\begin{algorithm}
\SetAlgoLined
\ \ Consider a physics-informed neural network $\bm{u}_{\bm{\theta}}(t, \bm{x})$ imposed the exact boundary conditions,
and the corresponding weighted loss function 
\begin{align}
  \mathcal{L}(\bm{\theta}) =\frac{1}{N_t} \sum_{i=0}^{N_t}  w_i \mathcal{L}(t_i, \bm{\theta}),
\end{align}
where $ \mathcal{L}(t_0, \bm{\theta}) =  \lambda_{ic} \mathcal{L}_{ic}(\bm{\theta})$ and for $1 \leq i \leq N_t$, $\mathcal{L}(t_i, \bm{\theta})$ is defined in Equation \ref{eq: temporal_residual}. All $w_i$ are initialized by 1. Then use $S$ steps of a gradient descent algorithm to update the parameters $\bm{\theta}$ as:

\For{$\epsilon = \epsilon_1, \dots, \epsilon_k$}
{
 \For{$n = 1, \dots, S$}
 {
  (a) Compute and update the temporal weights by
  \begin{align}
    w_i = \exp\left(- \epsilon \sum_{k=1}^{i-1}  \mathcal{L}(t_k, \bm{\theta})\right)    , \text{ for } i = 2, 3, \dots, N_t.
  \end{align}
 Here $\epsilon > 0$ is a user-defined hyper-parameter that determines the "slope" of temporal weights.
 
  (b) Update the parameters $\bm{\theta}$ via gradient descent
  \begin{align}
      \bm{\theta}_{n+1} = \bm{\theta}_{n} - \eta \nabla_{\bm{\theta}}\mathcal{L}(\bm{\theta}_n).
  \end{align}
   \If{$\min_i w_i > \delta$}
 {
 break
 }
}
}
\ \ The recommended hyper-parameters are $\lambda_{ic}=10^3$, $\delta = 0.99$ and $\{\epsilon_i\}_{i=1}^k = [10^{-2}, 10^{-1}, 10^{0}, 10^{1}, 10^{2}]$.
\caption{Causal training for physics-informed neural networks}
\label{alg}
\end{algorithm}

\paragraph{Connection to existing approaches:} It is worth pointing out that the proposed residual weighting strategy bears some similarity to the adaptive time sampling of Wight {\it et al.} \cite{wight2020solving}, since the effect of the  weights $w_i$ can be viewed as equivalent to changing the sampling density of collocation points. However, the method of Wight {\it et al.} has two main disadvantages in practice: a) the sampling density has to be manually designed for different problems and training iterations, and b)  an accurate approximation of the designed sampling density requires a large volume of collocation points, leading to a large computational cost. Besides, we remark that our method shares the same motivation with "time-marching" or "curriculum training" strategies \cite{wight2020solving, krishnapriyan2021characterizing, du2021evolutional, vadyala2022physics}, in the sense of respecting temporal causality by learning the solution sequentially within separate time-windows. In fact, our {\em causal training} strategy should not be viewed as a replacement of time-marching approaches, but instead as a crucial enhancement to those, given the fact that violations of causality may still occur within each time window of a time-marching algorithm.

\section{Practical considerations}\label{sec:practical}

As we will see in section \ref{sec:results}, high-order accuracy becomes a necessity for PINNs in order to tackle problems exhibiting sensitivity on initial data and strong spatio-temporal correlations (e.g. chaotic systems). Although PINNs are known for being incapable to achieve high-order accuracy in general, in this section we highlight a few extensions that can further enhance their performance in more challenging settings. Although these features are not deemed crucial for the successful application of Algorithm \ref{alg}, we have empirically observed that, for the problems considered in this work, they can lead to further enhancements in terms of accuracy and computational efficiency. 

\paragraph{Modified multi-layer perceptrons:}
\label{sec: modified_mlp}
In \cite{wang2021understanding} Wang {\em et al.} put forth a novel architecture that was demonstrated to outperform conventional MLPs across a variety of PINNs benchmarks. Here, we will refer to this architecture as "modified MLP". The forward pass of  a $L$-layer modified MLP is defined as follows
\begin{align}
    &\bm{U} = \sigma( \bm{X}  \bm{W}_1 + \bm{b}_1), \ \  \bm{V} = \sigma(\bm{X} \bm{W}_2  + \bm{b}_2), \\
    &\bm{H}^{(1)} = \sigma(\bm{X} \bm{W}^{(l)} + \bm{b}^{(l)}), \\
    &\bm{Z}^{(l)} = \sigma(\bm{H}^{(k)}\bm{W}^{(l)} + \bm{b}^{(l)}), \ \ l=1, \dots, L - 1,\\
    &\bm{H}^{(l+1)} = (1 - \bm{Z}^{(l)}) \odot \bm{U}  +  \bm{Z}^{(l)}  \odot \bm{V}, \ \  l=1, \dots, L - 1, \\
   & \bm{u}_{\bm{\theta}}(\bm{X}) = \bm{H}^{(L)}\bm{W}^{(L)}  + \bm{b}^{(L)},
\end{align}
where $\sigma$ denotes a nonlinear activation function, $\odot$ denotes a point-wise multiplication, and $\bm{X}$ denotes an batch of input coordinates. All trainable parameters are given by
\begin{align}
    \bm{\theta} = \{\bm{W}_1, \bm{b}_1, \bm{W}_2, \bm{b}_1, (\bm{W}^{(l)}, \bm{b}^{(l)})_{l=1}^L\}.
\end{align}
At first glance, this architecture seems to appear a bit complicated. However, notice that it is almost the same as a standard MLP  network, with the addition of two encoders and a minor modification in the forward pass.  Specifically, the inputs $\bm{X}$ are embedded into a feature space via two encoders $\bm{U}, \bm{V}$, respectively, and merged in each hidden layer of a standard MLP  using a point-wise multiplication. Based on our prior experience, the modified MLP architecture is shown to be more powerful than standard MLPs in terms of minimizing the PDE residuals and capturing sharp gradients \cite{wang2021understanding, wang2021eigenvector, wang2021learning, wang2021long}.

\paragraph{Exact periodic boundary conditions:}
\label{sec: PBC}
Recent work by Dong {\it et al.} \cite{dong2021method} showed how one can strictly impose periodic boundary conditions in PINNs as hard-constraints. We have empirically observed that this trick can simplify the training of PINNs and introduce some savings in terms of computational cost. To illustrate the main idea, let us consider enforcing periodic boundary conditions with period $P$ in a one-dimensional setting. To this end, we would like to make sure that a neural network returns periodic predictions as
\begin{align}\label{eq:periodic_constraint}
    u^{(l)}(a) = u^{(l)}(a + P), \quad l=0, 1, 2, \dots.
\end{align}
To enforce this constraint as part of the architecture itself, we construct a Fourier feature embedding of the form
\begin{align}
    \label{eq: 1D_Fourier}
    \bm{v}(x) = \left(1, \cos (\omega x), \sin (\omega x), \cos (2 \omega x), \sin (2 \omega x), \cdots, \cos (m \omega x), \sin (m \omega x) \right),
\end{align}
with $\omega = \frac{2 \pi}{L}$, and some non-negative integer $m$. Then, for any network representation $u_{\bm{\theta}}$, it can be proved that any $u_{\bm{\theta}}(v(x))$ exactly satisfies the periodic constraint of Equation \ref{eq:periodic_constraint} (see \cite{dong2021method} for a proof). 

The same idea can be extended to higher-dimensional domains. For instance, let $(x, y)$ denote the coordinates of a point in two dimensions, and suppose that $u(x, y)$ is a smooth periodic function to be approximated in a periodic cell $[a, a + P_x] \times [b, b + P_y]$, satisfying the following constraints 
\begin{align}
    &\frac{\partial^{l}}{\partial x^{l}} u\left(a, y\right)=\frac{\partial^{l}}{\partial x^{l}} u\left(a + P_x, y\right), \quad  y \in\left[b, b + P_y\right], \\
    &\frac{\partial^{l}}{\partial y^{l}} u\left(x, a\right)=\frac{\partial^{l}}{\partial y^{l}} u\left(x, b + P_y\right), \quad  x \in\left[a, a + P_x\right],
\end{align}
for $l=0, 1, 2, \dots$, where  $P_x$  and $P_y$ are the periods in the $x$ and $y$ directions, respectively. Similar to the one-dimensional setting, these constraints can be implicitly encoded in a neural network by constructing a two-dimensional Fourier features embedding as
\begin{align}
    \bm{v}(x, y) = \begin{bmatrix}
    \cos \left(\omega_{x} x\right) \cos \left(\omega_{y} y\right), \dots, \cos \left(n\omega_{x} x\right) \cos \left(m\omega_{y} y\right)\\
    \cos \left(\omega_{x} x\right) \sin \left(\omega_{y} y\right), \dots, \cos \left(n \omega_{x} x\right) \sin \left(m \omega_{y} y\right) \\
    \sin \left(\omega_{x} x\right) \cos \left(\omega_{y} y\right), \dots,  \sin \left(n \omega_{x} x\right) \cos \left(m \omega_{y} y\right) \\
    \sin \left(\omega_{x} x\right) \sin \left(\omega_{y} y\right), \dots, \sin \left(n \omega_{x} x\right) \sin \left(m \omega_{y} y\right)
    \end{bmatrix}
\end{align}
with $w_x = \frac{2 \pi}{P_x}, w_y = \frac{2 \pi}{P_y}$ and $m, n$ being some non-negative integers. Following \cite{dong2021method}, any network representation $u_{\bm{\theta}}(\bm{v}(x,y))$ is guaranteed to be periodic in the $x, y$ directions. 

For time-dependent problems, we simply concatenate the time coordinates $t$ with the constructed Fourier features embedding, i.e., $u_{\bm{\theta}}([t, \bm{v}(x)])$,  or $u_{\bm{\theta}}([t, \bm{v}(x, y)])$. Although in this work we will only consider periodic problems, other types of boundary conditions, including  Dirichlet, Neumann, Robin, etc., can also be enforced in a "hard" manner, see \cite{sukumar2021exact, lu2021physics} for more details.


\paragraph{Taylor-mode automatic differentiation for high-order derivatives:} 
\label{sec: taylor_mode}
Conventional forward- or reverse-mode automatic differentiation is known to incur a cost that scales exponentially -- both in terms of memory and computation -- with the order of differentiation. This can quickly introduce a bottleneck in cases where derivatives of order higher than two are required (see e.g. the Kuramoto-Sivashinsky benchmark considered in section \ref{sec:results}).
To address this drawback, here we employ Taylor-mode automatic differentiation \cite{griewank2008evaluating} in order to accelerate the computation of high-order derivatives. This is accomplished by leveraging a truncated Taylor polynomial approximation that allows for efficient computation of high-order derivatives of function compositions via the Fa\`a di Bruno formula \cite{griewank2008evaluating}
\begin{align}
    \frac{\partial^{n}}{\partial x_{1} \cdots \partial x_{n}} f(g(x))=\sum_{\sigma \in \pi_{\{1, \ldots, n\}}} f^{(|\sigma|)}(g(x)) \prod_{b \in \sigma} \frac{\partial^{|b|}}{\prod_{j \in b} \partial x_{j}} g(x),
\end{align}
where $\pi\{1, \dots ,n\}$ is the set of all partitions of the set $\{1, \dots, n\}$. It has been shown that Taylor-mode automatic differentiation enjoys much better scaling than conventional forward-mode or reverse-mode automatic differentiation, with its benefits becoming increasingly more dramatic as the order of differentiation is increased \cite{bettencourt2019taylor}. In terms of implementation, we leverage the \texttt{jax.jet} primitive accompanying the work of Bettencourt {\it et al.} \cite{bettencourt2019taylor, jax2018github}.

\paragraph{Parallel Training:}
Graphics processing units (GPUs) are the prevailing hardware choice for training PINNs, however these devices are often bound by their memory capacity. For more complex simulation scenarios (e.g. the Navier-Stokes benchmark in section \ref{sec:results}) we have empirically observed that using larger batch sizes during training leads to enhanced convergence and predictive accuracy. However, a desirable batch size might exceed the available memory that a single GPU can offer, therefore motivating the use of data-parallelism across multiple GPU devices. In order to facilitate this, we utilize synchronous data-parallelism across multiple GPUs, with each GPU storing an identical copy of all trainable parameters. In this paradigm, a batch of training data is split into sub-batches, one for each device. Specifically, batches of spatial and temporal points used to evaluate the training loss are generated randomly and independently on each available GPU, and gradients of the training loss are then aggregated across all devices with a collective reduce-mean operation. As such, each device can then update its own local copy of all trainable model parameters at each gradient descent iteration the using global gradient signal that is broadcasted across all devices. In our implementation, this is efficiently performed leveraging the \texttt{jax.pmap} primitive in JAX \cite{jax2018github}, allowing us to seamlessly scale our code to an arbitrary number of GPUs. The parallel performance of our implementation will be assessed via strong and weak scaling studies, as discussed in section \ref{sec:navier_stokes}.

\section{Results}\label{sec:results}


Our goal in this section is to demonstrate the effectiveness of the proposed {\em causal training} algorithm by providing  state-of-the-art numerical results for various types of differential equations exhibiting chaotic behavior, where existing PINNs formulations are destined for failure. Specifically, we will consider the forward simulation of the chaotic Lorenz system, the Kuramoto–Sivashinsky equation, and a two-dimensional simulation of decaying turbulence governed by the incompressible Navier-Stokes equations. Although these benchmarks can all be easily tackled using conventional numerical methods, they have remained elusive to PINNs since their initial conception \cite{psichogios1992hybrid, lagaris1998artificial}, and all the variants that followed the reincarnation of this framework by Raissi {et al.} \cite{raissi2019deep}.

Throughout all benchmarks, we will employ the modified MLP  architecture discussed in section \ref{sec:practical} equipped with hyperbolic tangent activation functions (Tanh) and initialized using the Glorot normal scheme \cite{glorot2010understanding}, unless otherwise stated. We will enforce periodic boundary conditions as hard constraints by constructing appropriate Fourier features embedding of the input, as discussed in section \ref{sec:practical}. All networks are trained via stochastic gradient descent using the Adam optimizer with default settings \cite{kingma2014adam} and an exponential learning rate decay with a decay-rate of 0.9 every $5,000$ training iterations.  As suggested by \cite{wight2020solving, krishnapriyan2021characterizing, mattey2022novel}, we will also employ  time-marching to reduce optimization difficulties. Specifically, we will split up the temporal domain of interest $[0, T]$ into sub-domains $[0, \Delta t], [\Delta t, 2 \Delta t], \dots [T-\Delta t, T]$, and train networks to learn the solution in each sub-domain, where the initial condition is obtained from the prediction of the previously trained network. At the end of training, the resulting PINN model can produce predictions for the target solution at any continuous query location in the global spatio-temporal domain.

All hyper-parameter settings, computational costs, implementation details and validation metrics are all discussed in the Appendix.  The code and data accompanying this manuscript will be made publicly available at \url{https://github.com/PredictiveIntelligenceLab/CausalPINNs}.

\subsection{Lorentz system}
As our first example, we consider the chaotic Lorenz system. It is well known that this system exhibits strong sensitivity to its initial conditions, which can trigger divergent trajectories in finite time if the numerical predictions sought are not sufficiently accurate. The system is described by the following ordinary differential equations
\begin{align}
    &\frac{\mathrm{d} x}{\mathrm{~d} t}=\sigma(y-x), \\
    &\frac{\mathrm{d} y}{\mathrm{~d} t}=x(\rho-z)-y, \\
    &\frac{\mathrm{d} z}{\mathrm{~d} t}=x y-\beta z.
\end{align}
These equations arise in studies of convection and instability in planetary atmospheric convection, where  $x$, $y$, and $z$ denote variables proportional to convective intensity, horizontal, and vertical temperature differences \cite{lorenz1963deterministic}.  Parameters $\rho, \sigma$ and $\beta$  denote the Prandtl number, Rayleigh number, and a geometric factor, respectively. The Lorenz system is well-known to be chaotic  for certain parameter values and initial conditions. Here, we consider a classical setting with $\sigma=3, \rho=28$, and $\beta=8/3$.  Our goal is to construct a PINNs model for learning the ODE solution up to time $T=20$, starting from an initial condition $[x(0), y(0), z(0)] = [1, 1, 1]$ that does not lie on the system's attractor. The employed PINN model architecture and training hyper-parameters are discussed in Appendix \ref{sec: parameters}.

Figure \ref{fig: lorentz_preds} shows the predicted trajectory against the reference trajectory obtained via a classical numerical solver (see Appendix \ref{sec: parameters} for more details), where an excellent agreement can be observed with a relative $L^2$ error $1.139e-02, 1.656e-02, 7.038e-03$  for the $x, y, z$ components, respectively. Moreover, all training losses are plotted in Appendix Figure \ref{fig: Lorentz_loss}. We can see that the stopping criterion $\min_i w_i > \delta$ discussed in section \ref{sec:causal_training} is satisfied for the training of each time window. It is worth pointing out that the proposed stopping criterion will not only benefit the predictive accuracy, but also save lots of computational costs. To verify this, we train the network by removing the stopping criterion and training for a fixed number of iterations for each time window under exactly the same hyper-parameter setting. Interestingly, as shown in Appendix \ref{fig: Lorentz_fixed_iterations_loss},  the training losses can achieve slightly lower values than the ones using the stopping criterion. However, the model predictions are less accurate, as some discrepancies can be clearly observed in Appendix Figure \ref{fig: lorentz_fixed_iterations_preds}. Although the reason behind this behavior still remains unclear, it appears that training the model for more iterations after the proposed stopping criterion has been met seems to give rise to over-fitting.

\begin{figure}
    \centering
    \includegraphics[width=0.8\textwidth]{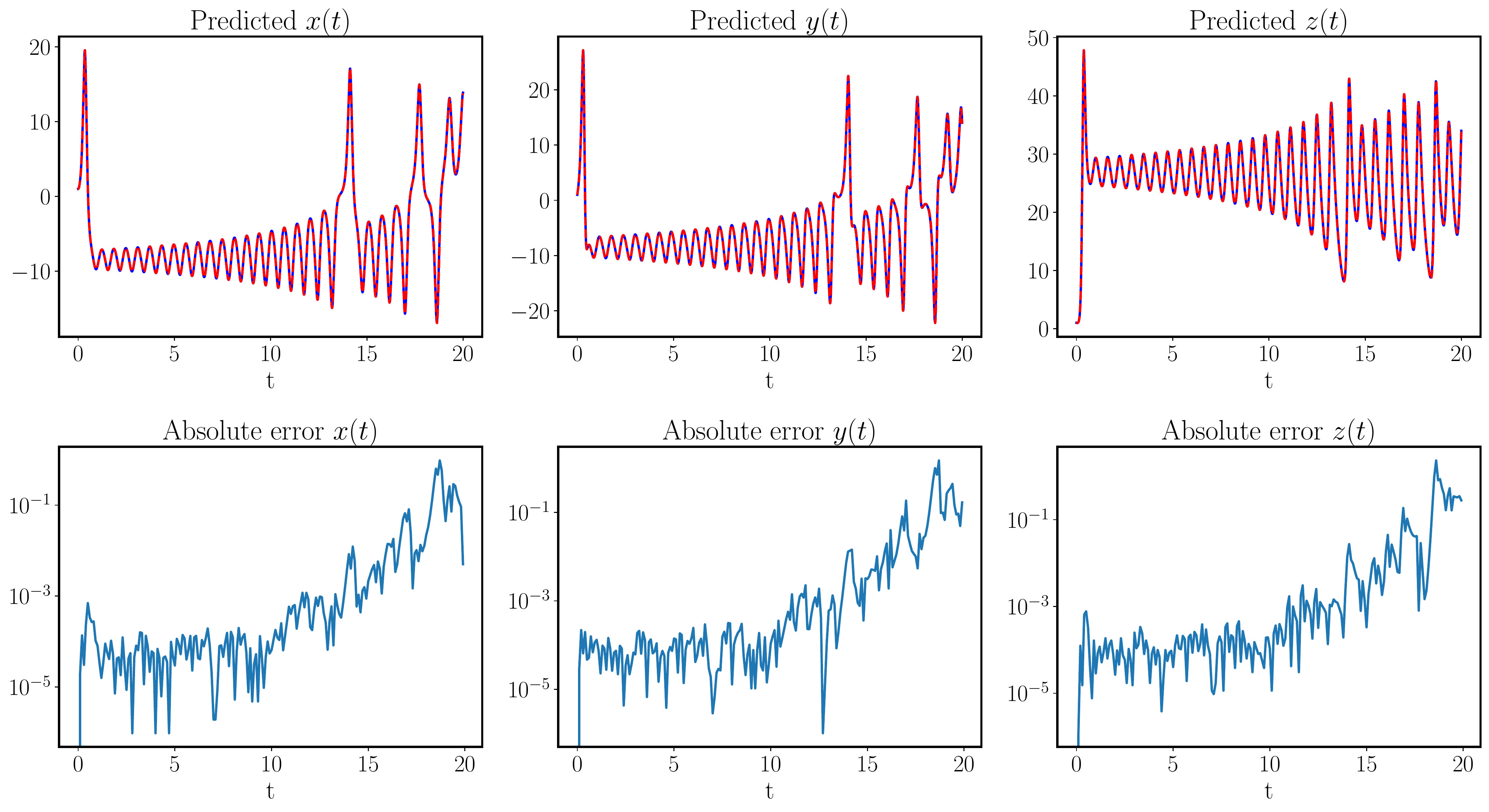}
    \caption{{\em Lorentz system:} Comparison between the predicted and reference solutions. }
    \label{fig: lorentz_preds}
\end{figure}

\subsection{Kuramoto–Sivashinsky equation}

The next example aims to illustrate the effectiveness of our method in tackling spatio-temporal chaotic systems. To this end, we consider one-dimensional Kuramoto–Sivashinsky equation, which has been independently derived in the context of reaction-diffusion systems \cite{kuramoto1976persistent} and flame front propagation \cite{sivashinsky1977nonlinear}. The Kuramoto–Sivashinsky equation exhibits a wealth of spatially and temporally nontrivial dynamical behavior including chaos, and has served as a model example in efforts to understand and predict the complex dynamical behavior associated with a variety of physical systems. The equation takes the form
\begin{align}
    \label{eq: ks}
    u_{t}+ \alpha u u_x + \beta  u_{x x}+ \gamma u_{x x x x}=0, 
\end{align}
subject to periodic boundary conditions and an initial condition 
\begin{align}
    u(0, x) = u_0(x).
\end{align}

\begin{figure}
     \centering
     \begin{subfigure}[b]{0.8\textwidth}
         \centering
         \includegraphics[width=\textwidth]{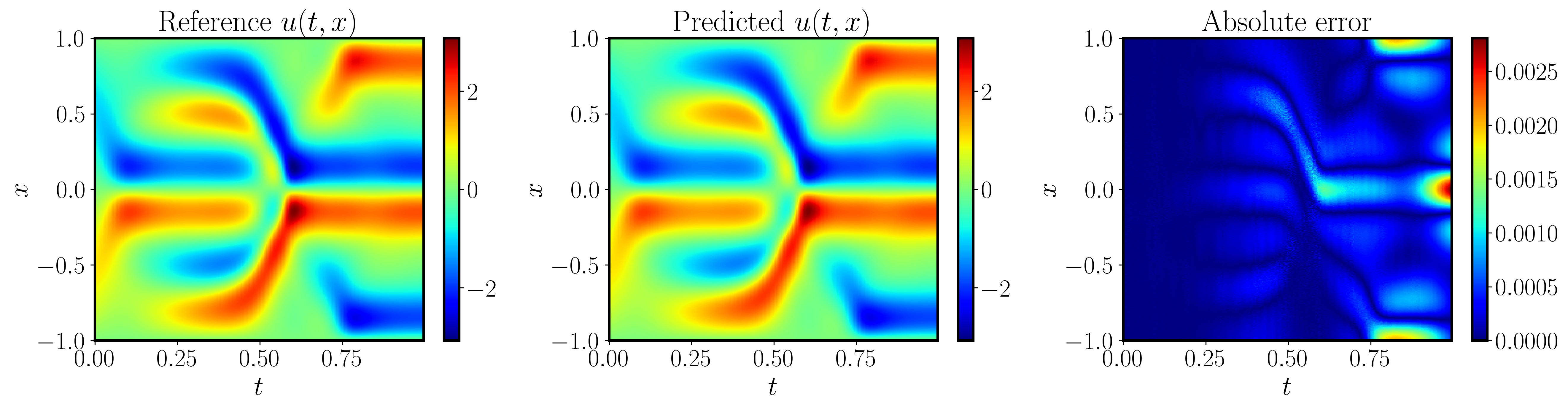}
     \end{subfigure}
     \begin{subfigure}[b]{0.7\textwidth}
         \centering
         \includegraphics[width=\textwidth]{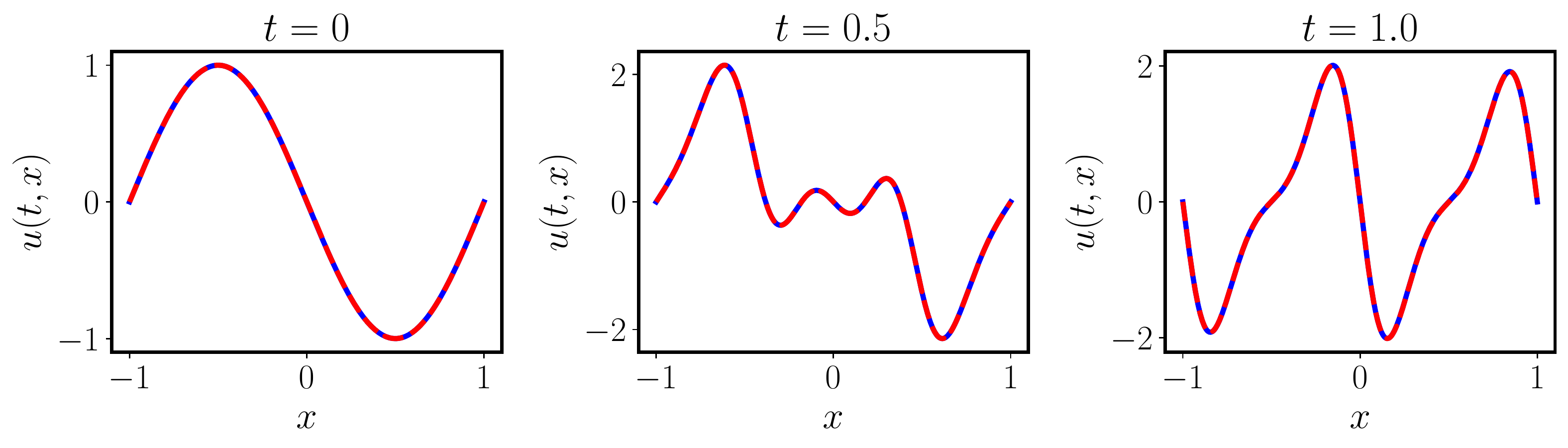}
     \end{subfigure}
        \caption{{\em Kuramoto–Sivashinsky equation (regular):} {\em Top:}  Reference solution versus the prediction of a trained physics-informed neural network using  Algorithm \ref{alg}. The resulting relative $L^2$ error is $3.49e-04$.  {\em Bottom:}  Comparison of the predicted and reference solutions corresponding to the three temporal snapshots at $t=0, 0.5, 1.0$. }
        \label{fig: KS_pred}
\end{figure}

\begin{figure}
     \centering
     \begin{subfigure}[b]{0.3\textwidth}
         \centering
         \includegraphics[width=\textwidth]{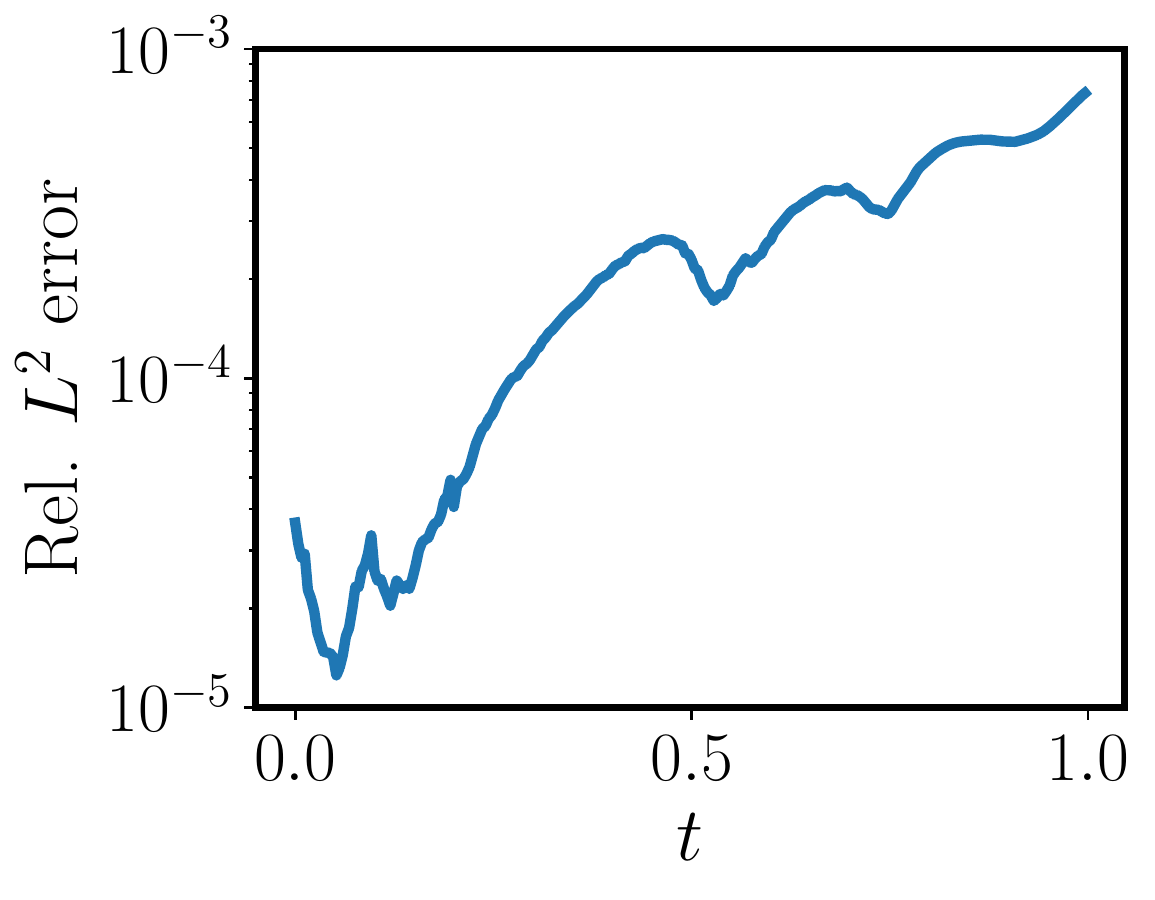}
     \end{subfigure}
     \begin{subfigure}[b]{0.3\textwidth}
         \centering
         \includegraphics[width=\textwidth]{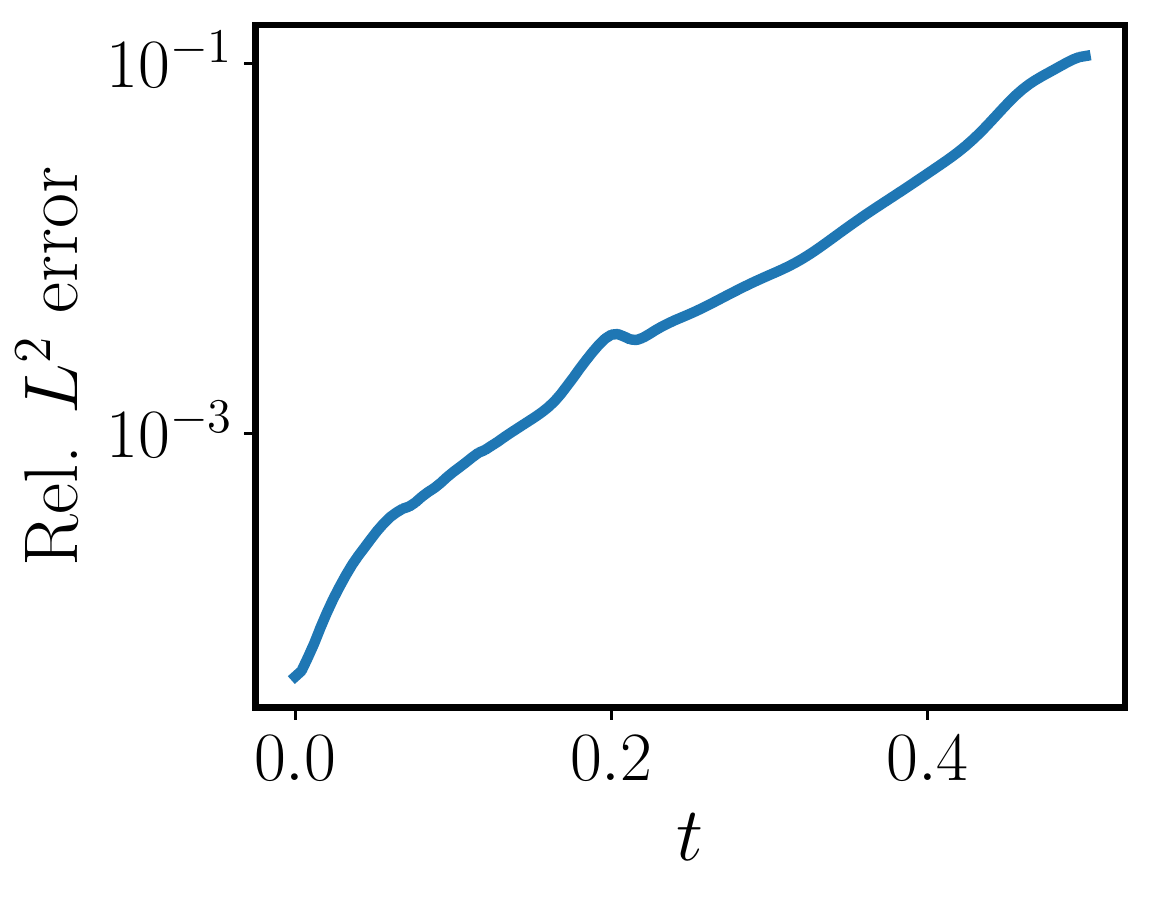}
     \end{subfigure}
        \caption{{\em Kuramoto–Sivashinsky equation:}  {\em Left:} Relative $L^2$ errors of Case I (regular).  {\em Right:} Relative $L^2$ errors of Case II (chaotic).}
        \label{fig: KS_error}
\end{figure}

\paragraph{Case I (regular):} We start with a relatively simple scenario by setting $\alpha = 5, \beta=0.5, \gamma=0.005$, and a spatial domain $[-1, 1]$. The initial condition is given by $u_0(x) = - \sin(\pi x)$. Our goal is to lean the associated solution up to time $T = 1$. A detailed visual assessment of the predicted solution is presented in Figure \ref{fig: KS_pred}. In particular, we present a comparison between the reference and the predicted solutions at different time instants $t=0, 0.5, 1.0$. It can be observed that the PINNs prediction achieves an excellent agreement with the reference solutions,  yielding an error of $3.49e-04$ measured in the relative $L^2$ norm. This is further illustrated by the temporal relative $L^2$ error shown in the left panel of Figure \ref{fig: KS_error}. Particularly, one may note that the error increases drastically by one order of magnitude for $t \in [0.4, 0.6]$ where the solution happens to experience a fast transition. This behavior is consistent with the larger loss values and the larger number of training iterations required before the stopping criterion is met, as observed in Appendix Figure \ref{fig: KS_loss}.

\begin{figure}
    \centering
    \includegraphics[width=0.7\textwidth]{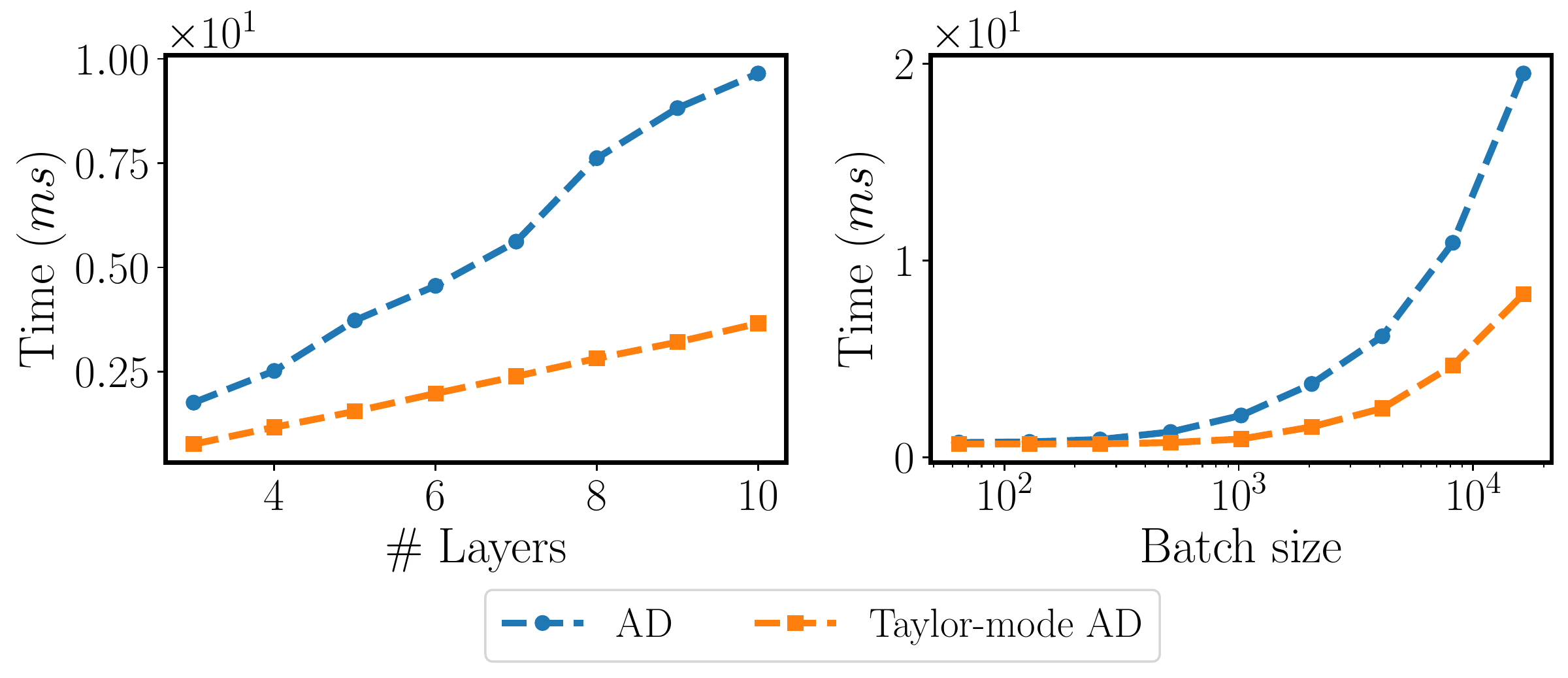}
    \caption{{\em Kuramoto–Sivashinsky equation (regular):} {\em Left:} Timing of evaluating the loss function of a PINN model with different number of layers. The rest hyper-parameters are the same as in Table \ref{tab: parameters}. {\em Right:} Timing of evaluating the forward pass of a PINN model with different batch sizes. The rest hyper-parameters are the same as in Table \ref{tab: parameters}.}
    \label{fig: KS_train_time}
\end{figure}

To highlight the computational efficiency of Taylor-mode automatic differentiation (Taylor-mode AD) discussed in section \ref{sec: taylor_mode}, here we provide a comparison in terms of computational cost against conventional reverse-mode automatic differentiation (AD) \cite{griewank2008evaluating}. Specifically, we consider PINN models with a different number of layers and batch sizes. As shown in Figure \ref{fig: KS_train_time}, Taylor-mode AD provides a significant advantage in terms of computational efficiency, allowing us to accommodate larger architectures and batch sizes. As a consequence, for the same architecture and batch size, we have consistently observed a speed-up of 3-5x in the total training time required for Taylor-mode AD versus conventional AD.

\begin{figure}
     \centering
     \begin{subfigure}[b]{0.8\textwidth}
         \centering
         \includegraphics[width=\textwidth]{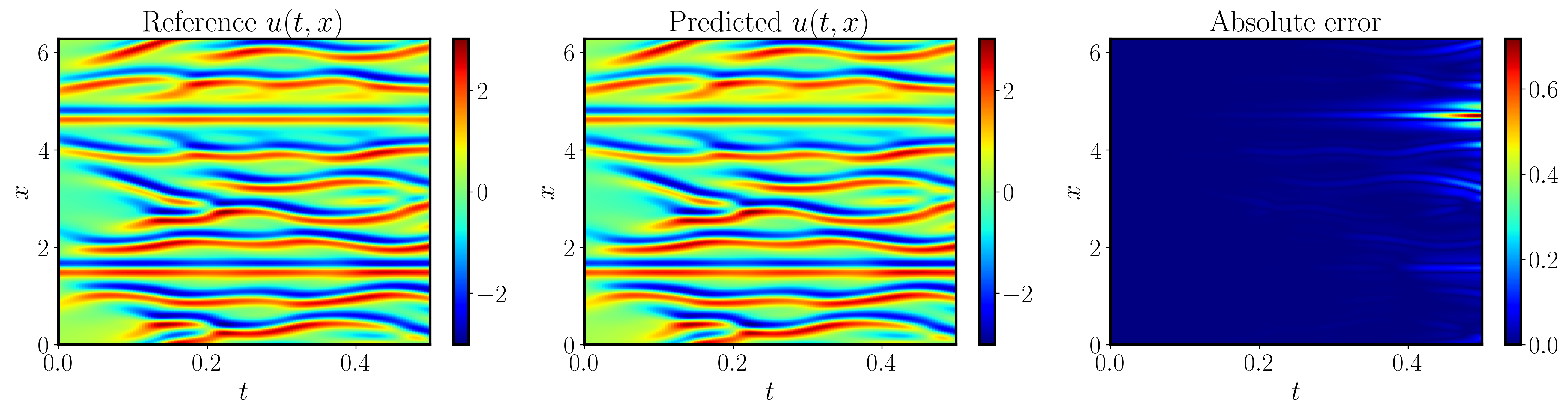}
     \end{subfigure}
     \begin{subfigure}[b]{0.7\textwidth}
         \centering
         \includegraphics[width=\textwidth]{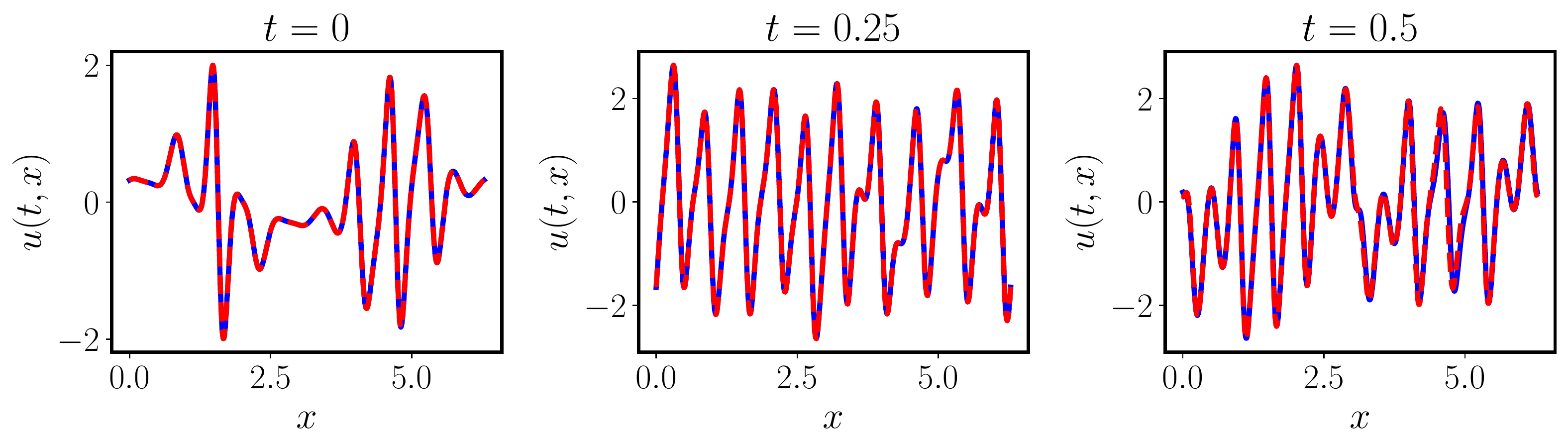}
     \end{subfigure}
        \caption{{\em Kuramoto–Sivashinsky equation (chaotic):} {\em Top:}  Reference solution versus the prediction of a trained physics-informed neural network using Algorithm \ref{alg}. The resulting relative $L^2$ error is $2.26e-02$.  {\em Bottom:}  Comparison of the predicted and reference solutions corresponding to the three temporal snapshots at $t=0, 0.25, 0.5$. }
        \label{fig: KS_chaotic_pred}
\end{figure}

\paragraph{Case II (chaotic):} We proceed by solving a more challenging case exhibiting  chaotic behavior, which remains stubbornly unsolved using existing PINNs formulations \cite{raissi2018deep}. 
Specifically, we set  $\alpha = 100/16, \beta=100/16^2,  \gamma=100/16^4$, for a fixed spatial domain in $[0, 2 \pi]$. Starting from an initial condition in the chaotic regime, we use PINNs to solve Kuramoto–Sivashinsky equation up to time $T = 0.5$. The results are summarized in Figure \ref{fig: KS_chaotic_pred}, from which one can see that the predicted solution is in good agreement with the reference solution obtained via classical spectral methods (see Appendix \ref{appendix: KS} for more details). The resulting relative $L^2$ error over the entire spatio-temporal domain is $2.46e-02$, which is visualized in the right panel of Figure \ref{fig: KS_error}. These results highly suggest that the proposed {\em causal training} algorithm enables PINN models to capture the intricate chaotic behavior of this system. 

\begin{figure}
     \centering
     \includegraphics[width=0.8\textwidth]{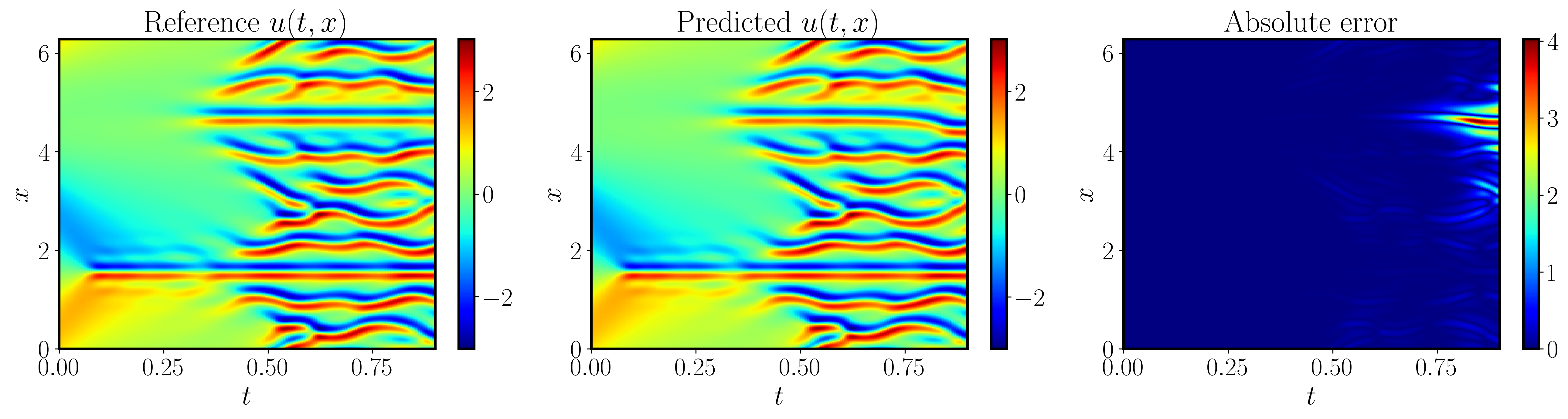}
     \caption{{\em Kuramoto–Sivashinsky equation (chaotic):} Reference solution versus the prediction of a trained physics-informed neural network using Algorithm \ref{alg}. The initial condition is $u_0(x) = \cos(x)(1 + \sin(x))$ An animation of the solution evolution is provided at \url{https://github.com/PredictiveIntelligenceLab/CausalPINNs\#kuramotosivashinsky-equation}.}
     \label{fig: KS_chaotic_full_pred}
 \end{figure}

From a critical standpoint, here we should also mention that difficulties can still arise in simulating the long-time behavior of chaotic systems. Figure \ref{fig: KS_chaotic_full_pred} summarizes our  results starting with a simple initial state $u_0(x) = \cos(x)(1 + \sin(x))$, and simulating the dynamics up to time $T = 0.9$. One can observe that the predicted solution accurately captures the transition to chaos at around $t=0.4$, while eventually loses accuracy after $t=0.8$ due to the chaotic nature of the problem and the inevitable numerical error accumulation of PINNs, leading to a relative $L^2$ error above $10\%$ for the final state. This highlights the crucial need for further enhancing the accuracy of PINN approximations in order to retain effectiveness in such complex regimes. Long-time integration in general, has been one PINNs' major drawbacks, and in future work we plan to address this via operator learning techniques as described in \cite{wang2021long}.


\subsection{Navier-Stokes equation}
\label{sec:navier_stokes}
To further emphasize the effectiveness of the proposed {\em causal training} algorithm for solving chaotic dynamical systems, in the last example, we consider a classical two-dimensional decaying turbulence example in a square domain with periodic boundary conditions. This problem can be modeled via the incompressible Navier-Stokes equations  expressed in the velocity-vorticity formulation
\begin{align}
w_t +\bm{u} \cdot \nabla w &= \frac{1}{\text{Re}} \Delta w,   \quad \text{ in }  [0, T] \times \Omega,  \\
\nabla \cdot \bm{u}  &=0,  \quad \text{ in }  [0, T] \times \Omega, \\
w(0, x, y) &=w_{0}(x, y),   \quad \text{ in }  \Omega,
\end{align}
where $\bm{u} = (u, v)$ denotes the flow velocity field, $w = \nabla \times \bm{u}$ denotes the vorticity, and $\text{Re}$ is the Reynolds number. In addition, we set $\Omega = [0, 2\pi]^2$ and $Re = 100$. Our goal is to use PINNs to simulate the flow up to $T = 1$.

\begin{figure}
    \centering
    \includegraphics[width=0.8\textwidth]{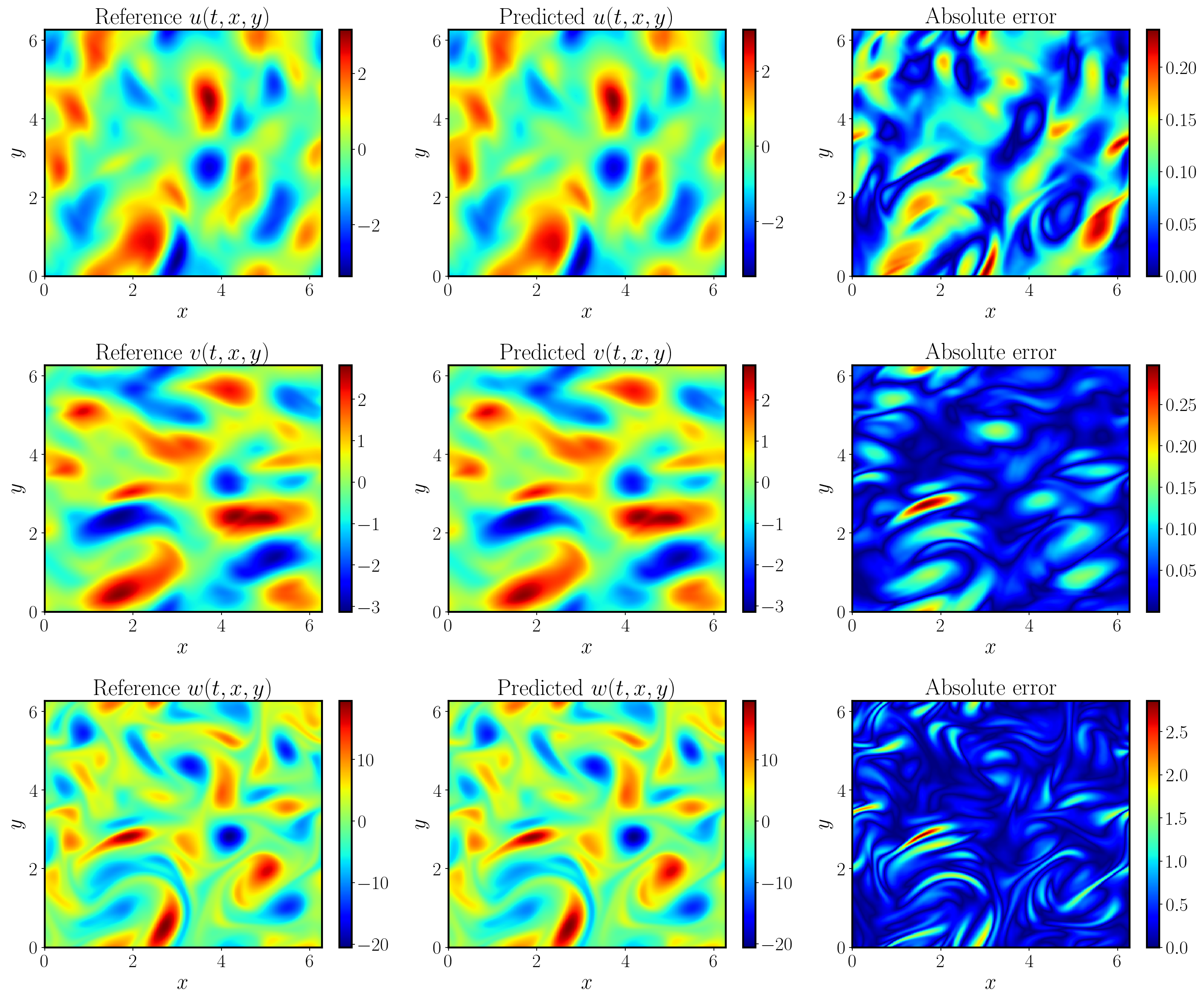}
    \caption{{\em Navier-Stokes equation:} Representative snapshot of the predicted velocity and vorticity versus the corresponding reference solution at $t = 1$. An animation of the solution evolution is provided at \url{https://github.com/PredictiveIntelligenceLab/CausalPINNs\#navier-stokes-equation}.}
    \label{fig: turbulence_preds_1}
\end{figure}

\begin{figure}
    \centering
    \includegraphics[width=0.8\textwidth]{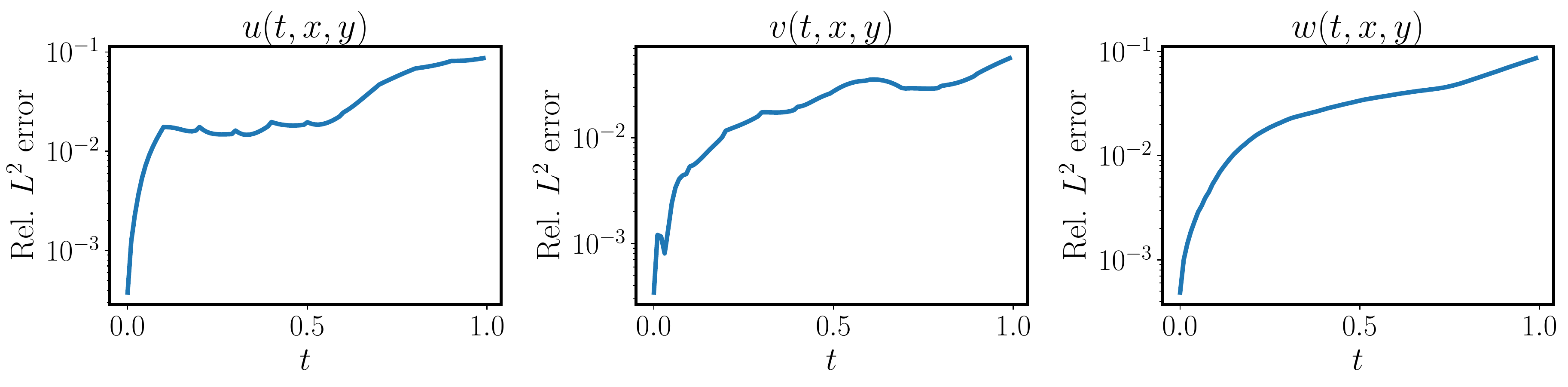}
    \caption{{\em Navier-Stokes equation:} Relative $L^2$ prediction errors for $u, v, w$, respectively.}
    \label{fig: turbulence_error}
\end{figure}

\begin{figure}
    \centering
    \includegraphics[width=0.8\textwidth]{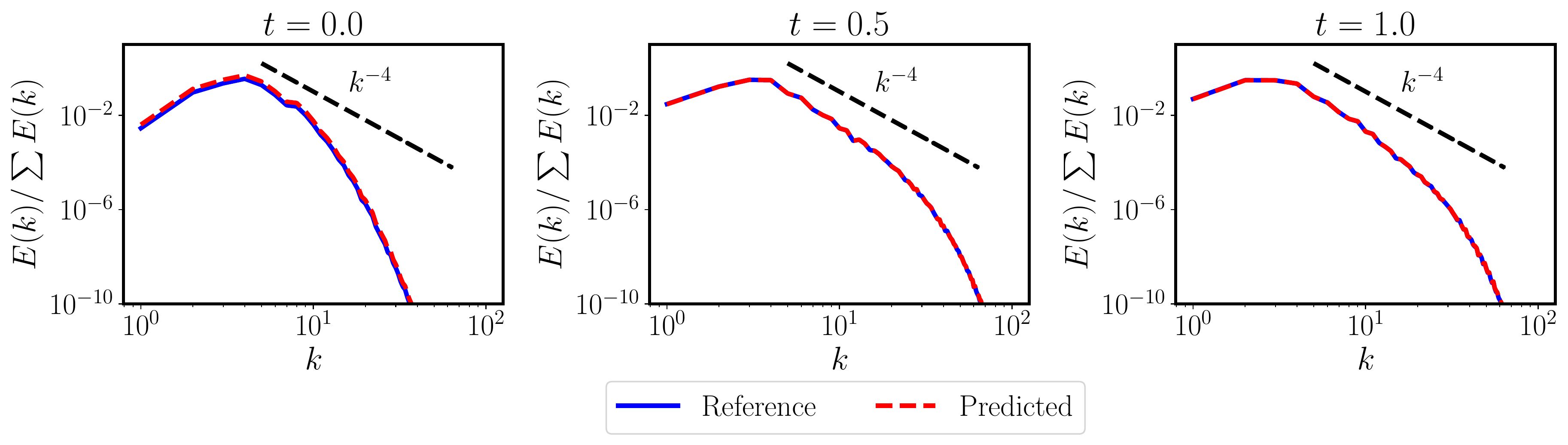}
    \caption{{\em Navier-Stokes equation:} Reference versus predicted normalized kinetic energy spectra at different time snapshots $t = 0.0, 0.5, 1.0$.}
    \label{fig: turbulence_TKE}
\end{figure}

Figure \ref{fig: turbulence_preds_1} presents the predicted velocity and vorticity field at $T = 1$. More detailed visual assessments are provided in Appendix \ref{appendix: NS}. We can see that all latent variables of interest are in good agreement with their corresponding reference solutions, yielding an error of $3.90e-02, 2.61e-02, 3.53e-02$ for $u,v,w$, respectively, over the entire spatio-temporal domain. This observation is further illustrated by the resulting errors reported in Figure \ref{fig: turbulence_error} and the computed energy spectrum in Figure \ref{fig: turbulence_TKE}. These results highlight the remarkable effectiveness of the proposed {\em causal training} algorithm, successfully enabling the PINNs model to capture such complicated turbulent flow without any training data.

\begin{figure}
    \centering
    \includegraphics[width=0.8\textwidth]{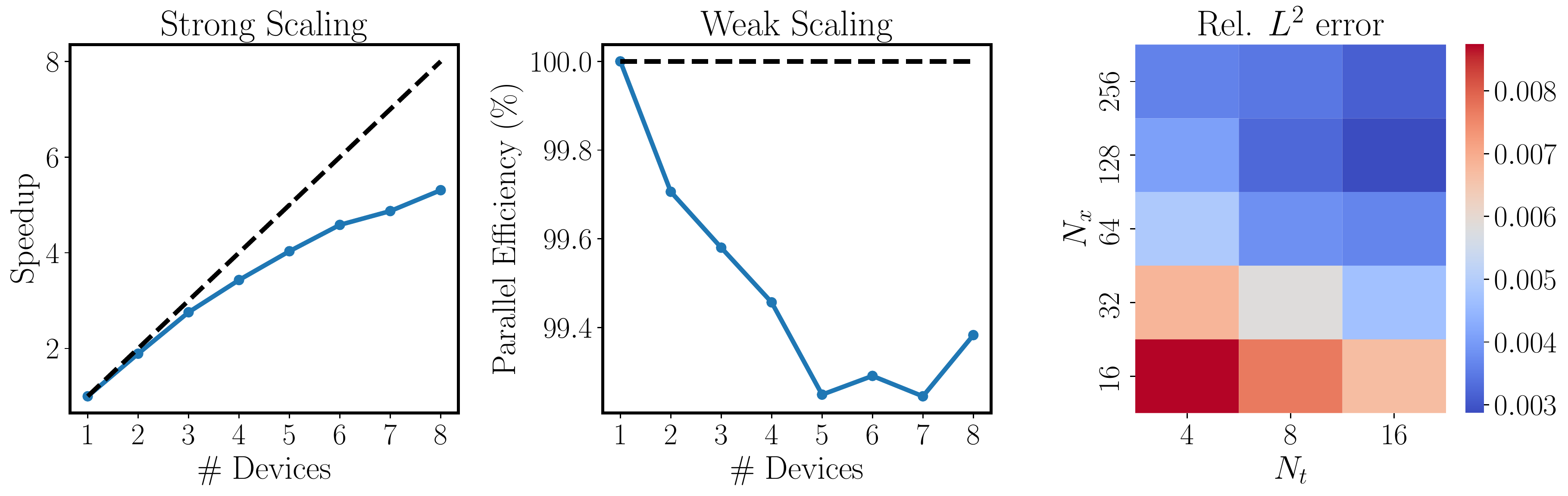}
    \caption{{\em Parallel Performance: Left: Strong Scaling:} Keeping the total-batch size for the problem fixed, we evaluate the speedup obtained when the batch is split across multiple devices. {\em Centre: Weak Scaling:} Keeping the batch-size on each GPU fixed, we report the efficiency of scaling by dividing the time taken on a single device over the time taken on $n$-devices. {\em Right: Effect of batch-size:} $L^2$ error for models trained till $t = 0.1$ using $N_t$ and $N_x$ points per iteration in the temporal and spatial domain respectively.}
    \label{fig: parallel_scaling}
\end{figure}

For this benchmark, we also report the performance of our parallel JAX implementation on a compute node equipped with 8 NVIDIA Ampere A6000 GPUs. We use an effective batch-size of $42,000$ spatio-temporal points sampled in each training iteration on each GPU with a network consisting of $6$ layers with $300$ neurons per layer. Figure \ref{fig: parallel_scaling} presents the scaling results obtained. To conduct a strong scaling study, we keep the problem size fixed and split the batch across several GPUs. As expected, we notice a speed-up, but the benefits deteriorate as the number of GPUs is increased beyond 4. We attribute this behavior to the fact that, for a fixed problem size, the compute load assigned to each GPU decreases as the number of devices is increased, leading to an under-utilization of each device. We have also performed a weak scaling study in which the number of points sampled per GPU is fixed. Under this setting, we report excellent parallel efficiency that remains above $99\%$ as the number of GPUs is increased. While we have only considered data-parallelism in this study, we may be able to obtain further speed-ups by considering a combination of data- and function-parallelism techniques \cite{schaarschmidt2021automap} in future studies. Figure \ref{fig: parallel_scaling} also reports the effect of batch-size of training on the resulting $L^2$ accuracy for the first time-window ($t\in[0,0.1]$). In general, we notice that an increase in batch-size results in higher accuracy of the network. This motivates the use of larger batch sizes through data-parallelism as a mechanism for enhancing the accuracy of PINNs in more challenging problems. 


\section{Discussion}\label{sec:discussion}

Physical systems possess an inherent causal structure that explains the fundamental relationship between causes and effects governing their dynamic evolution. In this work, we show that physics-informed neural networks are prone to violating that structure when trained to infer the solution of time-dependent PDEs. Specifically, by studying the limiting neural tangent kernel of PINNs we reveal an implicit bias indicating a preference of PINNs to first minimize PDE residuals at later times, before even fitting the initial data. We argue that this fundamental drawback is one of the key reasons why PINNs can fail in practice. To resolve this shortcoming, we propose a novel {\em causal training} algorithm that can restore physical causality during the training of a PINNs model by appropriately re-weighting the PDE residual loss at each iteration of gradient descent. Interestingly, this also leads to a simple stopping criterion for effectively assessing the convergence of the total training loss. We demonstrate that this simple modification alone is sufficient to achieve 10-100x improvements in accuracy compared to competing approaches, opening the path to tackling challenging problems that were not accessible to PINNs before, such as the chaotic Lorenz and Kuramoto-Sivashinsky equations, and the incompressible Navier-Stokes equations in the turbulent regime.

In this work we have solely focused on forward simulation problems, as we believe that these are the cases that most strongly expose the challenges and limitations in building PINNs models. While it is true that PINNs are currently better suited and have enjoyed far more success in tackling hybrid/inverse problems in which observational data is available, we believe that respecting causality is a crucial factor to consider when training a PINNs model, regardless of the forward/inverse nature of a given problem. To this end, in the inverse problem setting one should consider observational data as point sources of information, and ensure that PDE residuals are first adequately minimized at those locations before propagating information outwards. A more detailed exploration of this direction will be sought in future work.

We must also note that different problems are likely to pose a different causal structure. For example, in optimal control one needs to predict the state of a system by evolving its dynamics forward in time from a given initial condition, but also compute sensitivities with respect to a control input by evolving the adjoint system backwards in time from a given terminal condition that depends on the final system state. In this case, what we here refer to as ``temporal causality" takes a different form for the state (forward) and the co-state (adjoint) simulations. However, our main message remains the same: respecting causality matters, and training algorithms for PINNs should be designed to respect how information propagates according to the underlying principles that govern the evolution of a given system.

Given the rising prominence of PINNs across academic and industrial use cases, we consider this as a hallmark contribution that sets a new standard for what such models are capable of. We anticipate that the findings of this work will create new opportunities for the application of PINNs to  more complex scenarios across diverse domains, including fluid mechanics, electromagnetics, quantum mechanics, and elasticity. However, despite the encouraging results reported here, there is a still gap between the current progress in PINNs research and real-world applications. We have to admit that viewing PINNs as a forward PDE solver is significantly more time-consuming than the traditional numerical solvers.  Therefore, future research should focus on accelerating the training of PINNs. Distributed and parallel implementations can be of great help \cite{shukla2021parallel, jagtap2020extended} in this direction. Another aspect with great room for improvement is related to architecture design. Even though effective modifications such as the modified MLP discussed in section  \ref{sec:practical} and in \cite{wang2021understanding} can introduce noticeable gains in  accuracy, a niche architecture similar to what convolutional networks have been for vision or Transformers for language processing, is yet to be discovered for solving PDEs. To this end, we must recognize that training a PINN model is fundamentally different from solving conventional supervised learning tasks, requiring us to design more effective architectures for minimizing PDE residuals in a self-supervised manner. We believe that addressing these open questions will become an important piece of the puzzle in advancing the use of physics-informed machine learning as a reliable analysis tool in computational science and engineering.

\section*{Acknowledgements}
We would like to acknowledge support from the US Department of Energy under the Advanced Scientific Computing Research program (grant DE-SC0019116), the US Air Force (grant AFOSR FA9550-20-1-0060), and US Department of Energy/Advanced Research Projects Agency (grant DE-AR0001201). We also thank the developers of the software that enabled our research, including JAX \cite{jax2018github}, JAX-CFD\cite{Kochkov2021}, Matplotlib \cite{hunter2007matplotlib}, and NumPy \cite{harris2020array}.

\bibliographystyle{unsrt}
\bibliography{references}

\clearpage
\appendix
\input{appendix}

\end{document}

%% file: appendix.tex
\section{Nomenclature}

Table \ref{tab: Notations} summarizes the main symbols and notations used in this work.

\begin{table}[h]
\renewcommand{\arraystretch}{1.4}
    \centering
    \begin{tabular}{ll} 
    \Xhline{3\arrayrulewidth} 
    Notation     & Description \\
    \Xhline{3\arrayrulewidth} 
        PDE & Partial differential equation  \\
        PINN & Physics-informed neural network \\
        NTK & Neural Tangent Kernel \\
        $\bm{u}(\cdot)$ & solution of a  PDE \\
        $\mathcal{N}[\cdot]$         & a linear or non-linear differential operator \\
       $\mathcal{B}[\cdot]$ &  a boundary operator \\
        $u_{\bm{\theta}}(\cdot)$  &  neural network representation of the  latent PDE solution\\
        $\bm{\theta}$ &  all trainable parameters of a neural network \\
        $N_t$                & number of temporal collocation points \\
        $N_x$                & number of spatial collocation points \\
        $w_i$ &  residual weights at time $t_i$ \\
        $\epsilon$ & causality parameter    \\
        $\delta$ & stopping criterion threshold for terminating a training loop \\
      $\mathcal{L}_r(t, \bm{\theta})$ &  temporal residual loss \\
    $\mathcal{L}(\bm{\theta})$ &  aggregate training loss  \\
    \Xhline{3\arrayrulewidth}
    \end{tabular}
    \caption{{\em Nomenclature}: Summary of the main symbols and notations used in this work.}
    \label{tab: Notations}
\end{table}

\clearpage
\section{Hyper-Parameters}

\label{sec: parameters}

Table \ref{tab: parameters} summarizes the network hyper-parameters for all numerical experiments. We tuned these hyper-parameters manually, without attempting to find the absolute best hyper-parameter setting. This process can be automated in the future leveraging effective techniques for meta-learning and hyper-parameter optimization \cite{finn2017model}.

\begin{table}[h]
\renewcommand{\arraystretch}{1.2}
    \centering
    \begin{tabular}{c|c|ccccc}
        \Xhline{3\arrayrulewidth}
        Case   &  Architecture &  Depth &   Width & $N_t$ & $N_x$   \\
        \hline
        \multirow{2}{*}{ Allen-Cahn} & MLP  & 6 & 128  &  100 & 256  \\
        & Modified MLP  & 6 & 128  &  100 & 256  \\
        \hline
         Lorentz  & MLP & 5 & 512  &   256 & -  \\
         \hline
       Kuramoto–Sivashinsky (regular) & Modified MLP & 5 & 256  &   32 & 64  \\
       \hline
         Kuramoto–Sivashinsky (chaotic) & Modified MLP & 10 & 128  &   32 & 256  \\
         \hline
        Navier-Stokes & Modified MLP  & 6 & 128  & 64 & 512  \\
           \Xhline{3\arrayrulewidth}
    \end{tabular}
    \caption{Network architectures for each benchmark employed in this work. }
    \label{tab: parameters}
\end{table}

\section{Computational Cost}

{\bf Training:} Table \ref{tab: computational_cost} summarizes the computational cost of training PINNs.
The size of different models as well as network architectures are listed table \ref{tab: parameters}. All networks are trained using NVIDIA RTX A6000 graphics cards. 

\begin{table}[h]
\renewcommand{\arraystretch}{1.2}
    \centering
    \begin{tabular}{c|c|c|c|c}
     \Xhline{3\arrayrulewidth}
       Case   &  Architecture  &  \# Time windows &  Max. Iterations  & Training time (iter/sec)  \\
       \hline
    \multirow{2}{*}{ Allen-Cahn} & MLP  & 1 & $3 \times 10^5$  & 120.30
      \\ &Modified MLP  & 1  & $3 \times 10^5$  & 58.42  \\
      \hline
    Lorentz  &  MLP & 40 &  $1 \times 10^5$  & 957.41 \\
     \hline
     Kuramoto–Sivashinsky (regular) & Modified MLP & 10 &  $2 \times 10^5$  & 164.77 \\
      \hline
        Kuramoto–Sivashinsky (chaotic) & Modified MLP & 5 &  $2 \times 10^5$  &  28.22 \\
         \hline
      Navier-Stokes & Modified MLP & 10 &  $1 \times 10^5$  &  68.29\\    
    \Xhline{3\arrayrulewidth}
    \end{tabular}
    \caption{Computational cost reported timings are obtained on NVIDIA RTX A6000 graphics cards. We remark that "Max Iteration" is the maximum iteration for every tolerance $\epsilon$ in each time window. The default tolerance list is $[10^{-2}, 10^{-1}, 10^{0}, 10^{1}, 10^{2}]$ unless otherwise stated. The total number of iterations may vary for different examples due to the stopping criterion (see Algorithm \ref{alg}).}
    \label{tab: computational_cost}
\end{table}

\clearpage
\section{Allen-Cahn equation}
\label{appendix: AC}

\paragraph{Validation:}  We solve the Allen-Cahn equation  using conventional spectral methods. Specifically, assuming periodic boundary conditions, we start from the  initial condition $u_0(x) = x^2 \cos(\pi x)$ and integrate the system up to the final time $T=1$. Synthetic validation data are generated using the Chebfun package \cite{driscoll2014chebfun} with a spectral Fourier discretization with 512 modes and a fourth-order stiff time-stepping scheme (ETDRK4) \cite{cox2002exponential} with time-step size $10^{-5}$.

\begin{figure}[h]
     \centering
     \begin{subfigure}[b]{0.8\textwidth}
         \centering
         \includegraphics[width=\textwidth]{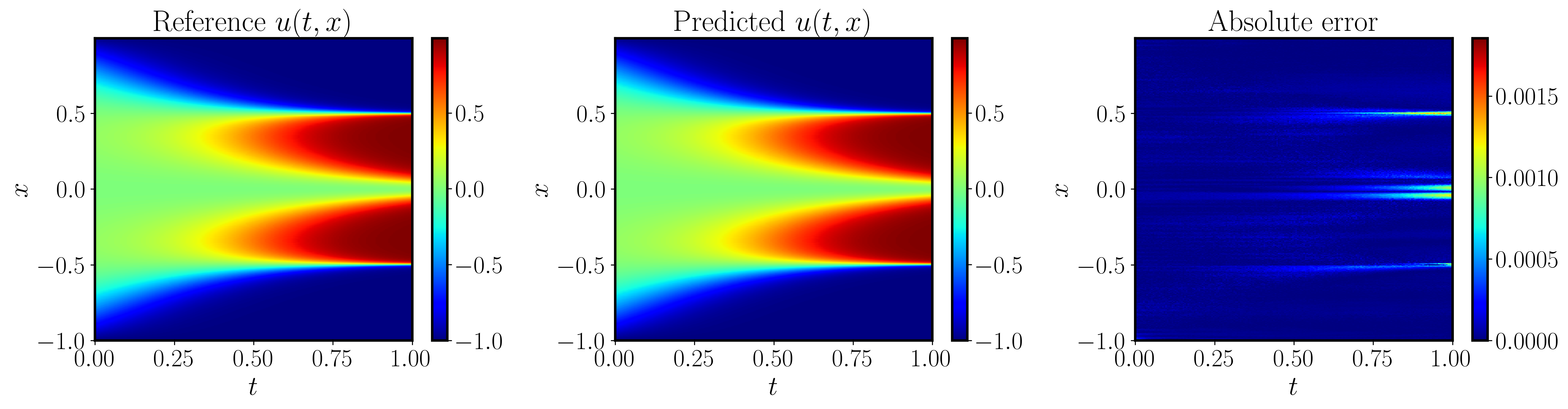}
     \end{subfigure}
     \begin{subfigure}[b]{0.7\textwidth}
         \centering
         \includegraphics[width=\textwidth]{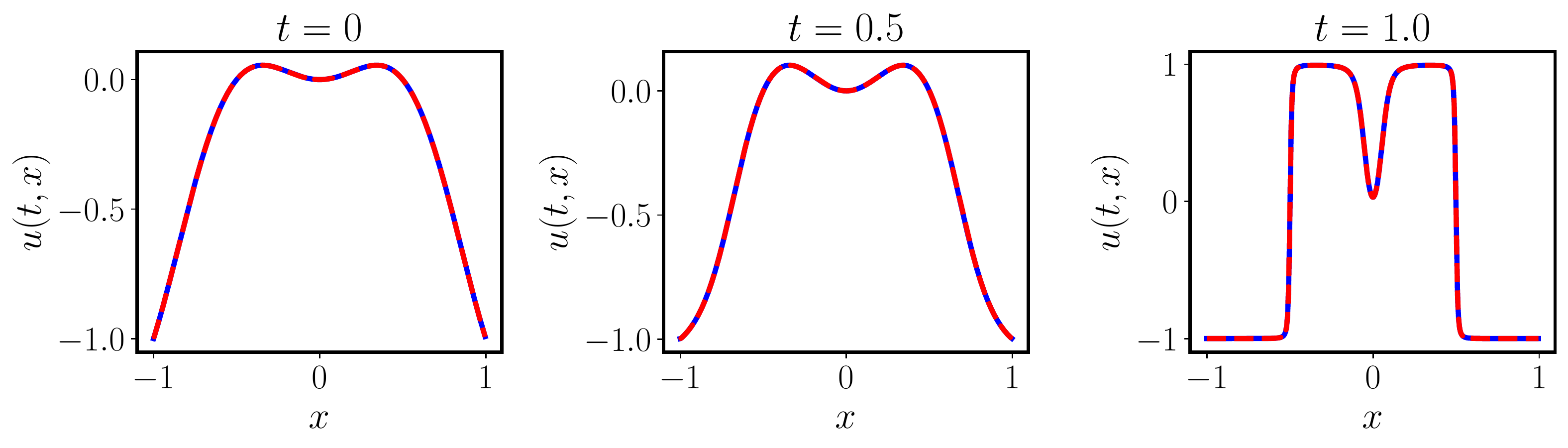}
     \end{subfigure}
        \caption{{\em Allen-Cahn equation:} {\em Top:}  Exact solution versus the prediction of a trained physics-informed neural network using Algorithm \ref{alg} and modified MLP. The resulting relative $L^2$ error is $2.46e-04$.  {\em Bottom:}  Comparison of the predicted and exact solutions corresponding to the three temporal snapshots at $t=0.0, 0.5, 1.0$. }
        \label{fig: AC_TW_modified_MLP_PINN_pred}
\end{figure}

\begin{figure}[h]
    \centering
    \includegraphics[width=0.9\textwidth]{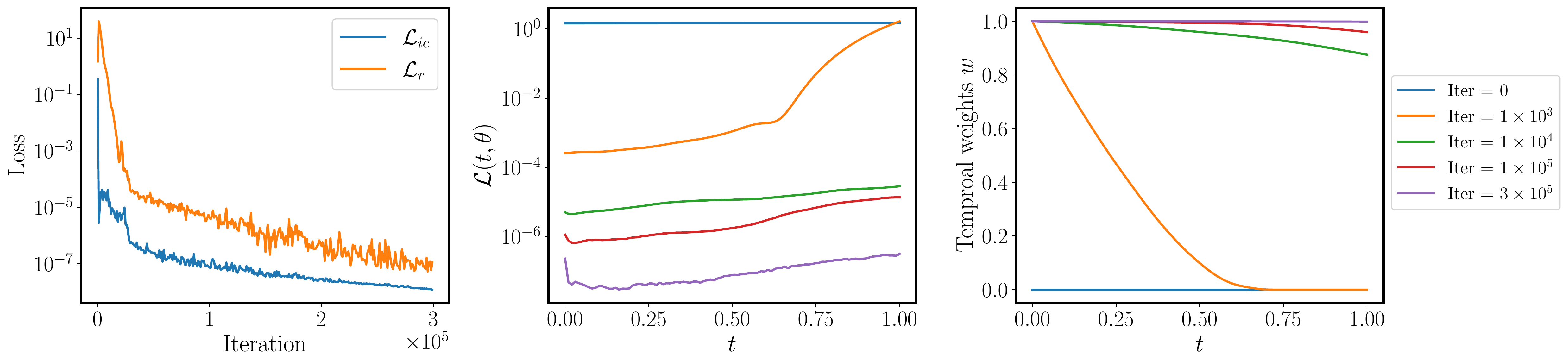}
       \caption{{\em Allen-Cahn equation:} {\em Left:} Loss convergence of training a physics-informed neural network using  Algorithm \ref{alg}.  {\em Middle:} Temporal residual loss $\mathcal{L}(t, \bm{\theta})$ at different training iteration. {\em Right:} Temporal weights at different training iteration.}
    \label{fig: AC_TW_modified_MLP_PINN_loss_L_t_weights}
\end{figure}

\clearpage
\section{Lorentz system}

\paragraph{Validation:} The reference solution is obtained using \texttt{scipy.integrate.odeint} with default settings.

\paragraph{PINNs implementation:} We split the whole domain $[0, 20]$ into 40 disjoint time windows of size $\Delta t = 0.5$. For each time window, we proceed by representing the latent variables of interest by a 5-layer fully-connected neural network  $\bm{u}_{\bm{\theta}}$ with 512 neurons per hidden layer
\begin{align}
    t \xrightarrow{\bm{u}_{\bm{\theta}}} [x_{\bm{\theta}}, y_{\bm{\theta}}, z_{\bm{\theta}}].
\end{align}
Since Lorentz system is highly sensitive to the initial condition, we exactly impose the initial condition by 
\begin{align}
    &\hat{x}_{\bm{\theta}}(t) = x_{\bm{\theta}}(t) \cdot t + x(0),  \\
    &\hat{y}_{\bm{\theta}}(t) = y_{\bm{\theta}}(t) \cdot t + y(0), \\
    & \hat{z}_{\bm{\theta}}(t) = z_{\bm{\theta}}(t) \cdot t + z(0).\end{align}
Then the loss function can be reduced to the residual loss
\begin{align}
    \mathcal{L}_r(\bm{\theta}) &=   \frac{1}{N_t} \sum_{i=1}^{N_t} w_i \left|\frac{\mathrm{d} \hat{x}_{\bm{\theta}}}{\mathrm{~d} t}(t_i) - \sigma \left(\hat{y}_{\bm{\theta}}(t_i)-\hat{x}_{\bm{\theta}}(t_i)\right) \right|\\
    & + \frac{1}{N_t} \sum_{i=1}^{N_t} w_i \left|\frac{\mathrm{d} \hat{y}_{\bm{\theta}}}{\mathrm{~d} t}(t_i) -\hat{x}_{\bm{\theta}}(t_i)(\rho-\hat{z}_{\bm{\theta}}(t_i))- \hat{y}_{\bm{\theta}}(t_i)  \right|\\
    & + \frac{1}{N_t} \sum_{i=1}^{N_t} w_i \left| \frac{\mathrm{d} \hat{z}_{\bm{\theta}}}{\mathrm{~d} t}(t_i) - \hat{x}_{\bm{\theta}}(t_i) \hat{y}_{\bm{\theta}}(t_i) + \beta \hat{z}_{\bm{\theta}}(t_i)    \right|,
\end{align}
where  $\{t_i\}_{i=1}^{N_t}$ is a uniform grid in $[0, \Delta t]$.  For this example, we set $N_t = 256$ and train the network with full-batch gradient descent. The temporal weights are updated by the proposed algorithm.

\begin{figure}[h]
    \centering
    \includegraphics[width=0.8\textwidth]{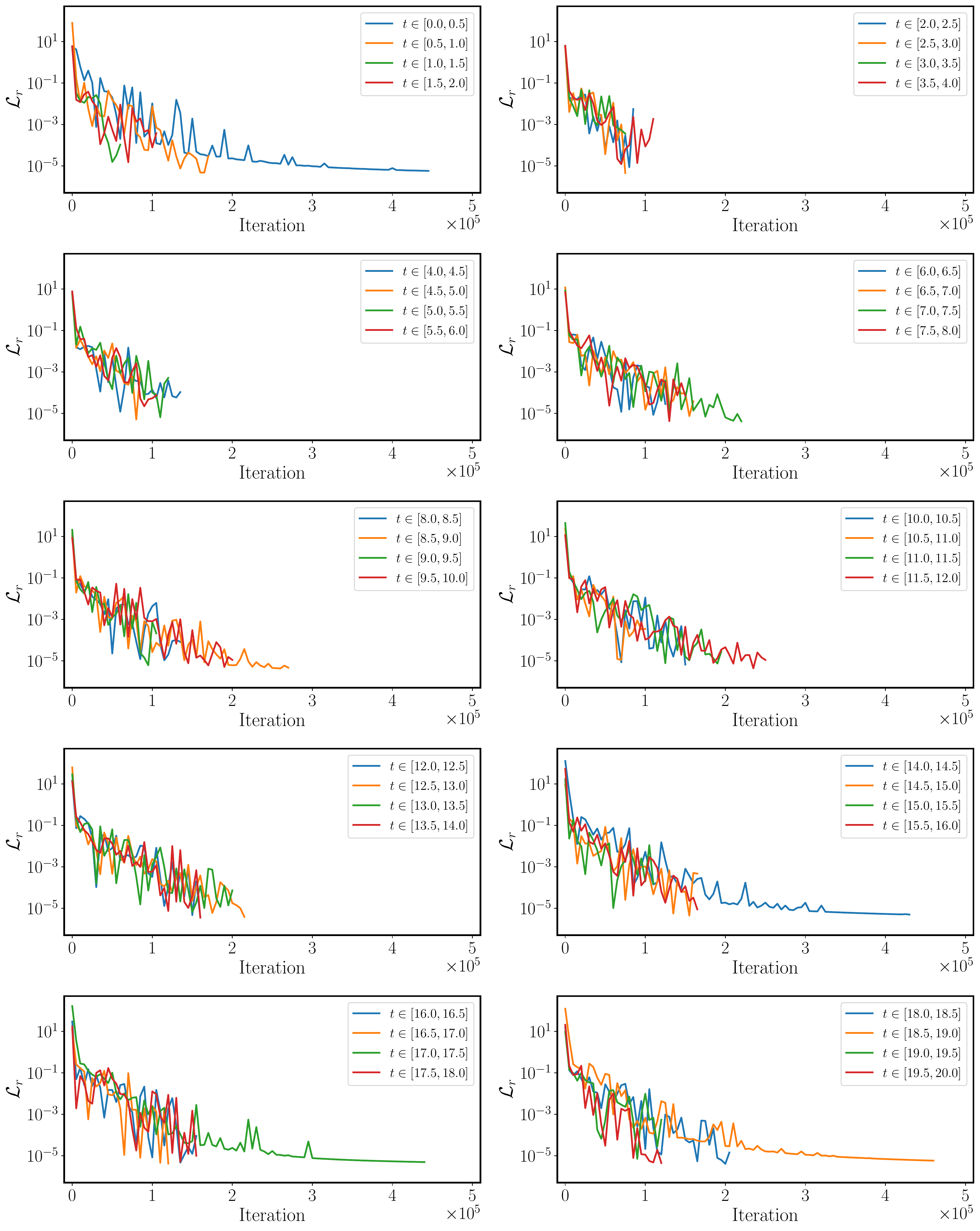}
    \caption{{\em Lorentz system:}  {\em Left:} Loss convergence of training a physics-informed neural network using  Algorithm \ref{alg} for every time window.  }
    \label{fig: Lorentz_loss}
\end{figure}

\begin{figure}[h]
    \centering
    \includegraphics[width=0.8\textwidth]{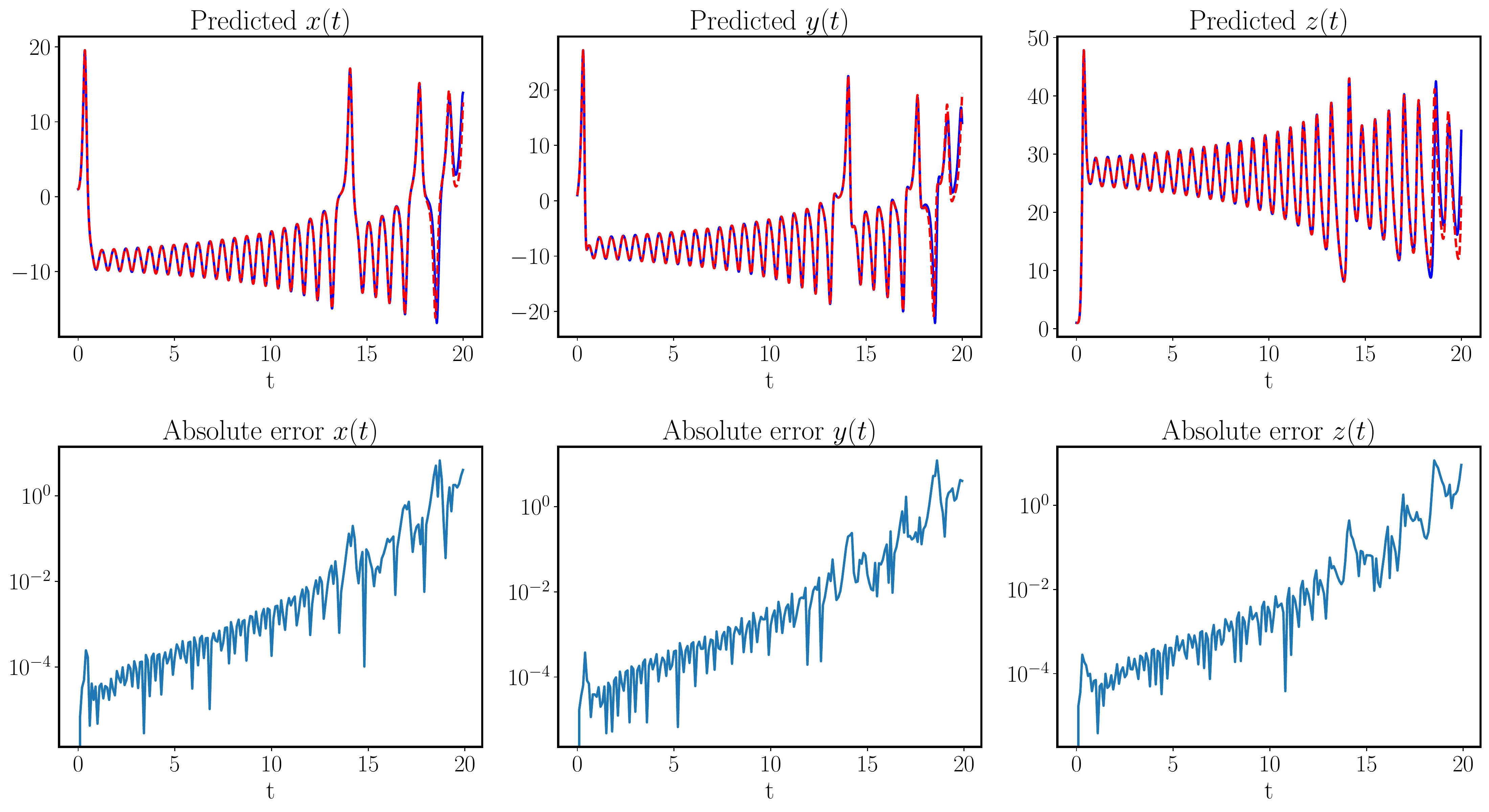}
    \caption{{\em Lorentz system:} Reference solutions versus the predicted solutions obtained by training a physics-informed neural network using  Algorithm \ref{alg} with fixed iterations. }
    \label{fig: lorentz_fixed_iterations_preds}
\end{figure}

\begin{figure}[h]
    \centering
    \includegraphics[width=0.8\textwidth]{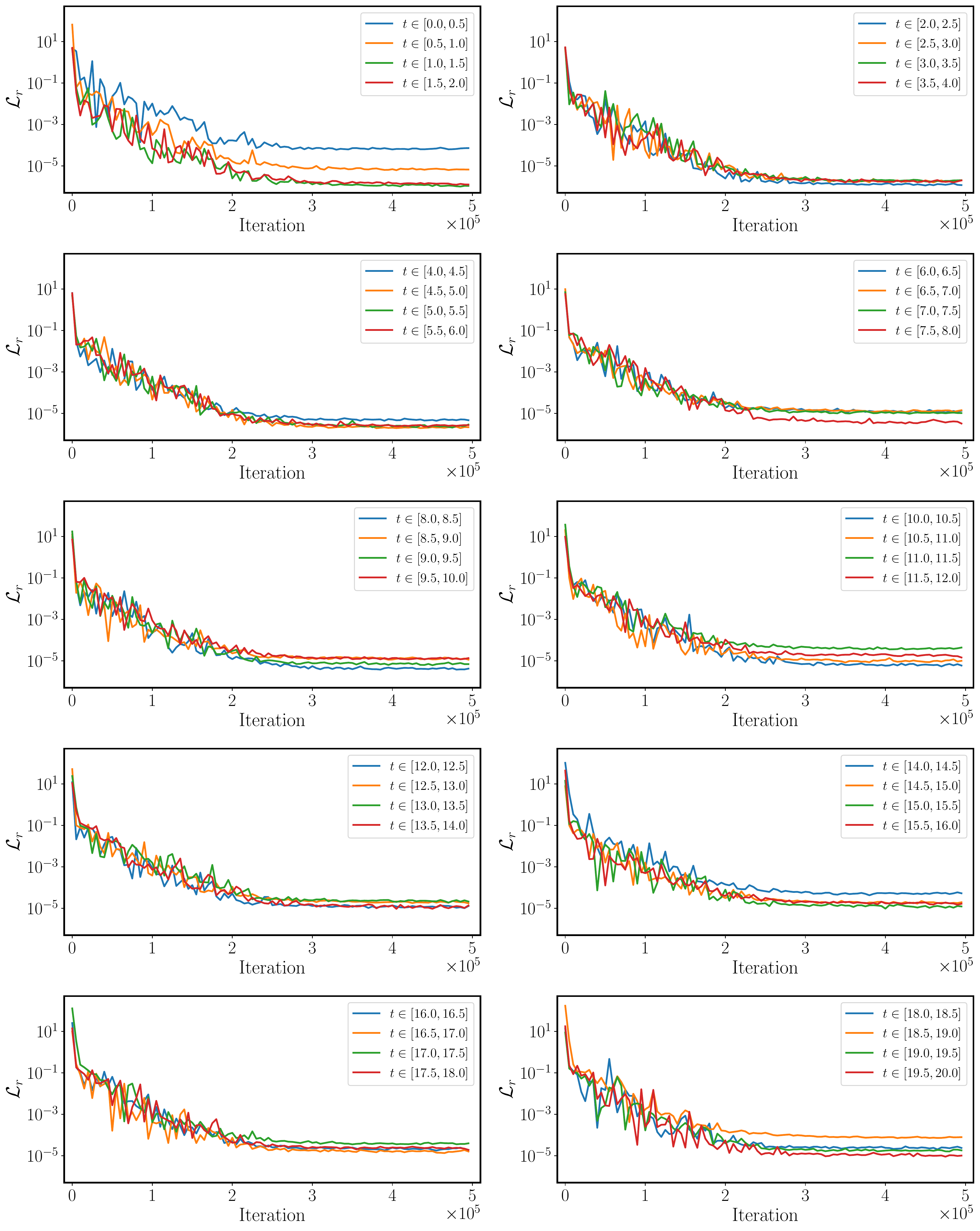}
    \caption{{\em Lorentz system:} {\em Left:} Loss convergence of training a physics-informed neural network using Algorithm \ref{alg} for every time window.}
    \label{fig: Lorentz_fixed_iterations_loss}
\end{figure}

\clearpage
\section{Kuramoto–Sivashinsky equation}
\label{appendix: KS}

\paragraph{Validation:} For case I (regular), we solve the Kuramoto–Sivashinsky equation  using conventional spectral methods. Specifically, assuming  periodic boundary conditions, we start from the  initial condition $u_0(x) = - \sin(\pi x)$ and integrate the Equation \ref{eq: ks} up to the final time $T=1$. Synthetic validation data are generated using the Chebfun package \cite{driscoll2014chebfun} with a spectral Fourier discretization with 512 modes and a fourth-order stiff time-stepping scheme (ETDRK4) \cite{cox2002exponential} with time-step size $10^{-5}$. For case II (chaotic), we perform the same procedure with the initial condition $u_0(x) = \cos(x) (1 + \sin(x))$. Then we select the numerical solution at $t=0.5$ as our initial condition for the PINNs simulation.

\paragraph{PINNs implementation:}  For Case I (regular), we split the temporal domain $[0, 1]$ into 10 time windows of size $\Delta t= 0.1$. Then we approximate the solution of each time window by a 5-layer modified MLP $u_{\bm{\theta}}$ with 256 neurons per hidden layer and encoded periodicity. It allows us to define the PDE residual by
\begin{align}
    \mathcal{R}[u_{\bm{\theta}}] =  \frac{\partial  u_{\bm{\theta}}}{\partial t}+ \alpha    u_{\bm{\theta}} \frac{\partial  u_{\bm{\theta}}}{\partial x} + \beta \frac{\partial^2 u_{\bm{\theta}}}{\partial x^2} + \gamma \frac{\partial^4 u_{\bm{\theta}}}{\partial x^4}.
\end{align}
Then, we can formulate the following loss function
\begin{align}
    \mathcal{L}(\bm{\theta}) =  \frac{1}{N_t}\sum_{i=0}^{N_t} w_i \mathcal{L}(t_i, \bm{\theta}),
\end{align}
where
\begin{align}
    \mathcal{L}(t_0, \bm{\theta}) &= \lambda_{ic} \frac{1}{N_x}\sum_{j=1}^{N_x} \left| u_{\bm{\theta}}(0, x_j) - u_0(x_j)  \right|^2, \\
   \mathcal{L}(t_i, \bm{\theta}) &= \frac{1}{N_x}\sum_{j=1}^{N_x} \left|  \mathcal{R}[u_{\bm{\theta}}](t_i, x_j) \right|^2, \text{ for } 1 \leq i \leq N_t.
\end{align}
Here we set $N_t = 32, N_x = 64$ and $\{t_i\}_{i=1}^{N_t}, \{x_j\}_{j=1}^{N_x}$ are randomly sampled in $[0, \Delta t]$ and $[-1, 1]$, respectively at each iteration of gradient descent. Particularly, we take $\lambda_{ic} = 10^3$ for better enforcing the initial condition. The network is trained by minimizing the above loss function via mini-batch gradient descent using the proposed algorithm.

 For Case II (chaotic): We split the temporal domain $[0, 0.5]$ into 5 time windows of size $\Delta t = 0.1$. Then we perform the same procedure except for employing a 10-layer  modified MLP with 128 neurons per hidden layer and setting $\lambda_{ic} = 10^4$.

\paragraph{Remark:} For both cases, we employ Taylor-mode automatic differentiation \cite{bettencourt2019taylor}  to accelerate the computation of high-order derivatives (see section \ref{sec:practical}).

\begin{figure}[h]
     \centering
     \begin{subfigure}[b]{0.9\textwidth}
         \centering
         \includegraphics[width=\textwidth]{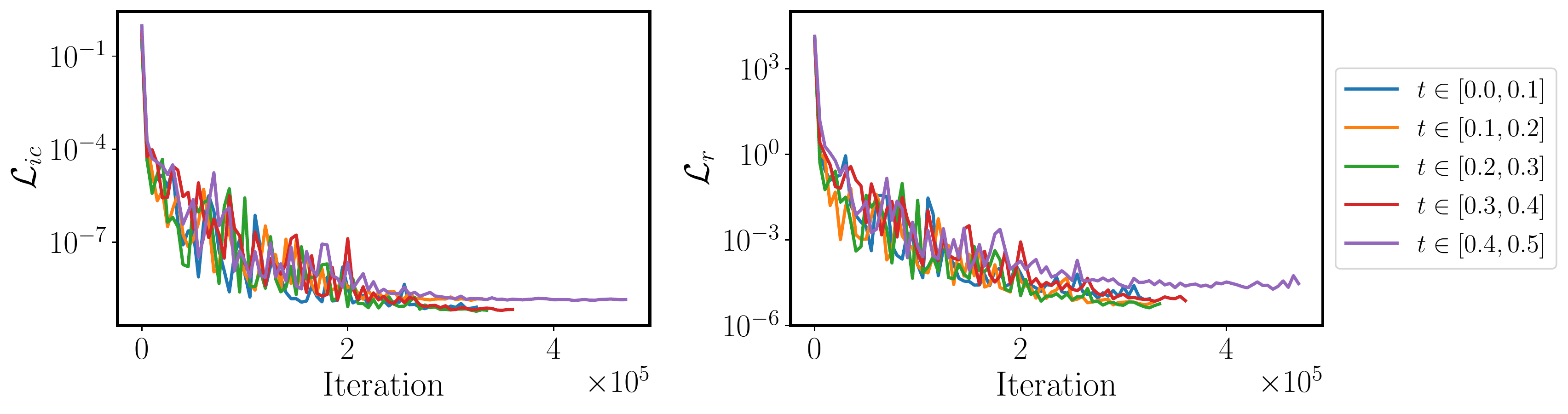}
     \end{subfigure}
     \begin{subfigure}[b]{0.9\textwidth}
         \centering
         \includegraphics[width=\textwidth]{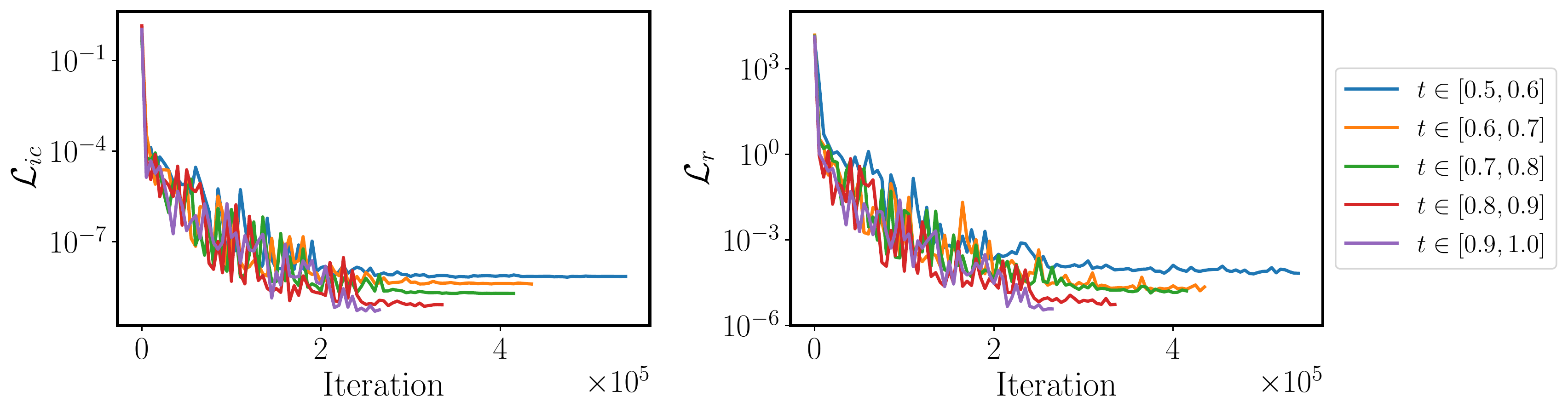}
     \end{subfigure}
        \caption{{\em Kuramoto–Sivashinsky equation (regular):} Loss convergence of training a physics-informed neural network using  Algorithm \ref{alg} for every time window. }
        \label{fig: KS_loss}
\end{figure}

 \begin{figure}[h]
     \centering
     \begin{subfigure}[b]{0.9\textwidth}
         \centering
         \includegraphics[width=\textwidth]{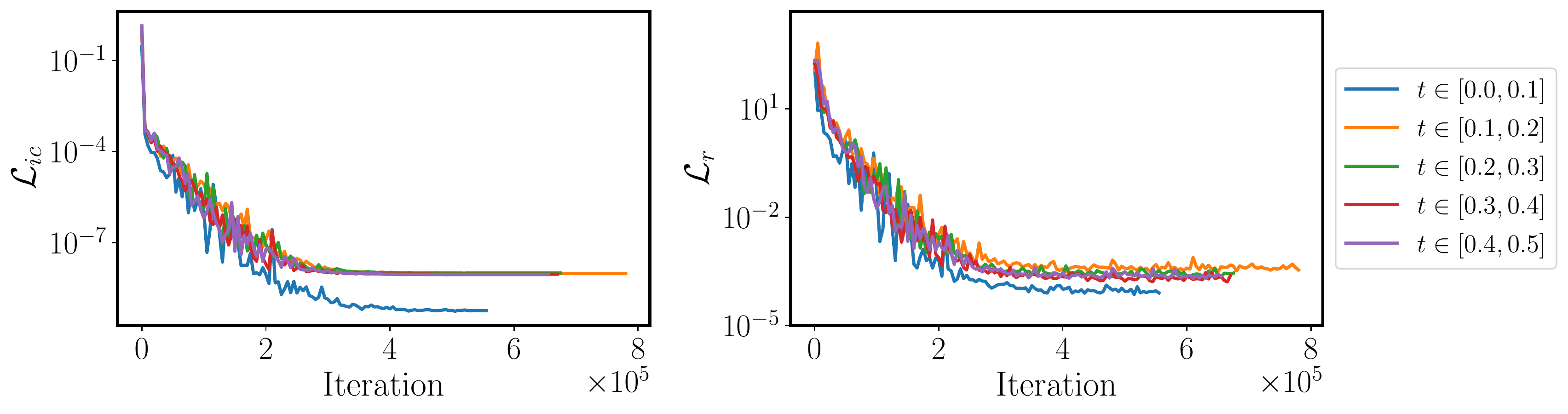}
     \end{subfigure}
        \caption{{\em Kuramoto–Sivashinsky equation (chaotic):} Loss convergence of training a physics-informed neural network using Algorithm \ref{alg} for every time window.}
        \label{fig: KS_chatoic_loss}
\end{figure}

\clearpage
\section{Navier-Stokes equation}
\label{appendix: NS}

\paragraph{Validation:} We simulate two-dimensional decaying turbulence in a periodic box using the JAX-CFD \cite{Kochkov2021} incompressible Navier-Stokes solver. A high-resolution validation data-set is created  by simulating an initial divergence free velocity field with the given maximum velocity $v_{\max} = 5$. The flow is solved using a Fourier spectral collocation method on a $1024 \times 1024$ uniform mesh with a time step of $dt = 10^{-4}$ \cite{Kochkov2021}.

\paragraph{PINNs implementation:} Similar to the previous examples, the time domain $[0, 1]$ is decomposed into 10 time windows of size $\Delta t = 0.1$. We proceed by representing the velocity field by a 6-layer modified MLP with 128 neurons per hidden layer
\begin{align}
    [t, x, y] \xrightarrow{\bm{u}_{\bm{\theta}}} [u_{\bm{\theta}}, v_{\bm{\theta}}].
\end{align}
Then the vorticity  can be approximated  by $w_{\bm{\theta}} = \partial_x  v_{\bm{\theta}} - \partial_y u_{\bm{\theta}} $ using automatic differentiation. Now we can define the PDE residual by
\begin{align}
    \mathcal{R}_{\bm{\theta}}^{w} &=  \frac{\partial w_{\bm{\theta}}}{\partial t} + u_{\bm{\theta}}  \frac{\partial w_{\bm{\theta}}}{\partial x} +  v_{\bm{\theta}}  \frac{\partial w_{\bm{\theta}}}{\partial y} - \frac{1}{\text{Re}} (   \frac{\partial^2 w_{\bm{\theta}}}{\partial x^2}  +  \frac{\partial^2 w_{\bm{\theta}}}{\partial y^2} ), \\
    \mathcal{R}_{\bm{\theta}}^{c} &= \frac{\partial u_{\bm{\theta}}}{\partial x} + \frac{\partial v_{\bm{\theta}}}{\partial y}.
\end{align}
It allows to define the loss function by
\begin{align}
    \mathcal{L}(\bm{\theta}) = \frac{1}{N_t} \sum_{i=0}^{N_t} w_i \mathcal{L}(t_i, \bm{\theta}), 
\end{align}
where
\begin{align}
    \mathcal{L}(t_0,\bm{\theta}) = \frac{\lambda_{ic}}{N_x} &\sum_{j=1}^{N_x}  \left| u_{\bm{\theta}}(0, x_j, y_j) - u_0(0, x_j, y_j) \right|^2 \\
    &+ \left| v_{\bm{\theta}}(0, x_j, y_j) - v_0(0, x_j, y_j) \right|^2 
    \\
    &+ \left| w_{\bm{\theta}}(0, x_j, y_j) - w_0(0, x_j, y_j) \right|^2 
\end{align}
and
\begin{align}
     \mathcal{L}(t_i,\bm{\theta}) &=  \frac{\lambda_{w}}{N_x} \sum_{j=1}^{N_x}  \left|  \mathcal{R}_{\bm{\theta}}^{w}(t_i, x_j, y_j) \right|^2 
     + \frac{\lambda_{c}}{N_x} \sum_{j=1}^{N_x}  \left|  \mathcal{R}_{\bm{\theta}}^{c}(t_i, x_j, y_j) \right|^2, \text{ for } 1 \leq i \leq N_t. 
\end{align}
For this example we set $N_t = 64, N_x = 512$ and $\lambda_w = 1, \lambda_c = 10^2, \lambda_{ic}=10^4$. The temporal and spatial collocation points are randomly sampled from $[0, 1]$ and $[0, 2\pi]^2$, respectively. It is worth noting that we also enforce the initial velocity field $(u_0, v_0)$ as additional constraints for better convergence. This is not a severe restriction since the velocity field can be obtained from the vorticity by solving the associated Poisson's equation or from the network representation directly. 

\begin{figure}[h]
    \centering
    \includegraphics[width=0.8\textwidth]{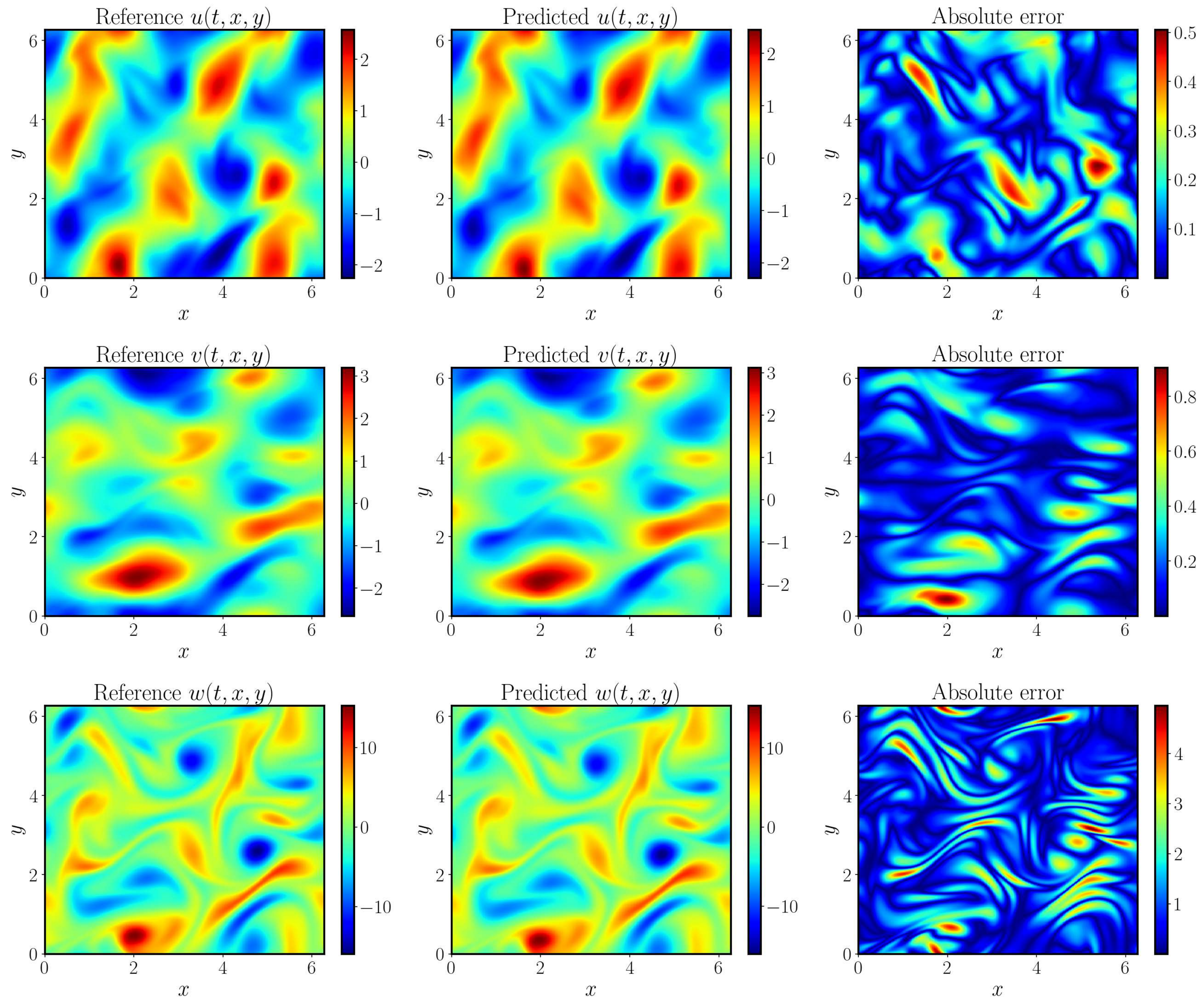}
    \caption{{\em Navier-Stokes:} Predicted $u,v,w$ against the ground truth at $t=2$.}
    \label{fig: turbulence_preds_full}
\end{figure}
\begin{figure}[h]
    \centering
    \includegraphics[width=0.8\textwidth]{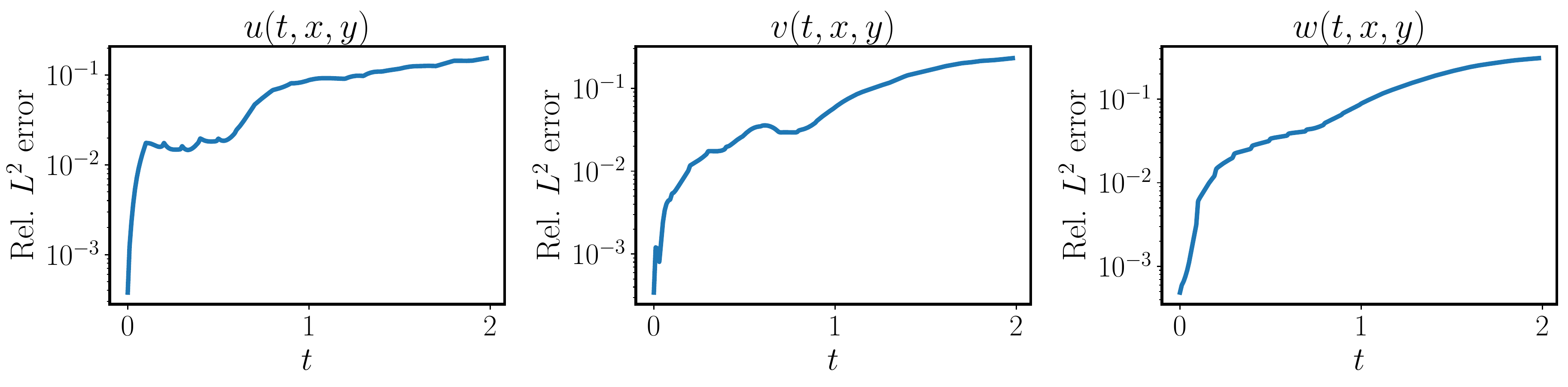}
    \caption{{\em Navier-Stokes:} Relative $L^2$ errors of $u,v,w$, respectively.}
    \label{fig: turbulence_error_full}
\end{figure}

Furthermore, in Appendix we also present our results simulating the turbulent flow up to $T=2$. Figure \ref{fig: turbulence_preds_full} presents the visualizations of the predicted velocity and vorticity field at the final state. The predictive accuracy is quantified in Figure \ref{fig: turbulence_error_full}. Although the resulting relative $L^2$ error is above $10\%$, our model predictions seem to be qualitatively correct against the corresponding ground truth.  

\begin{figure}[h]
    \centering
    \includegraphics[width=0.8\textwidth]{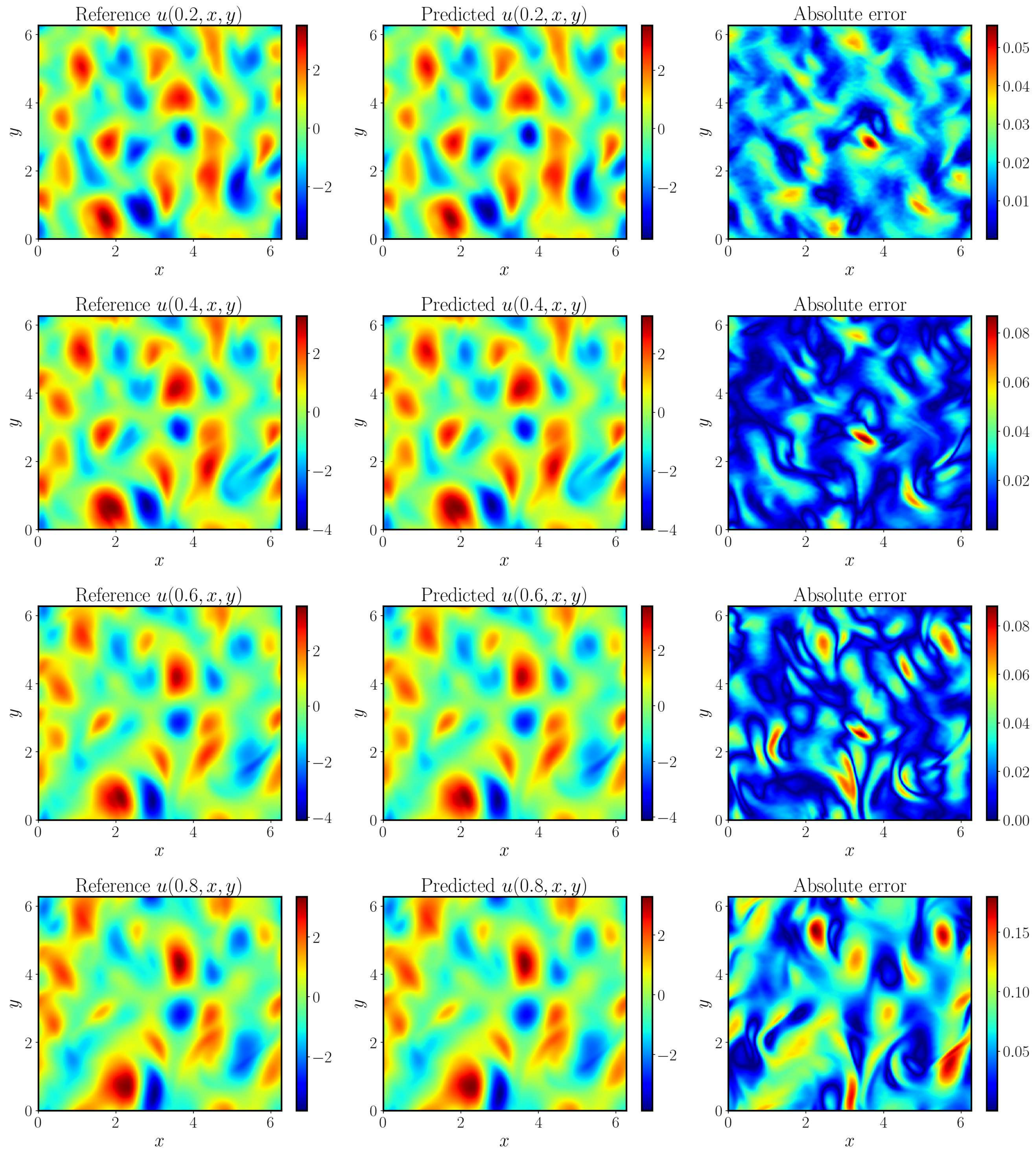}
    \caption{{\em Navier-Stokes:} Representative snapshots of the predicted $u$ against the ground truth at $t=0.2, 0.4, 0.6, 0.8$.}
    \label{fig: turbulence_u_preds}
\end{figure}

\begin{figure}[h]
    \centering
    \includegraphics[width=0.8\textwidth]{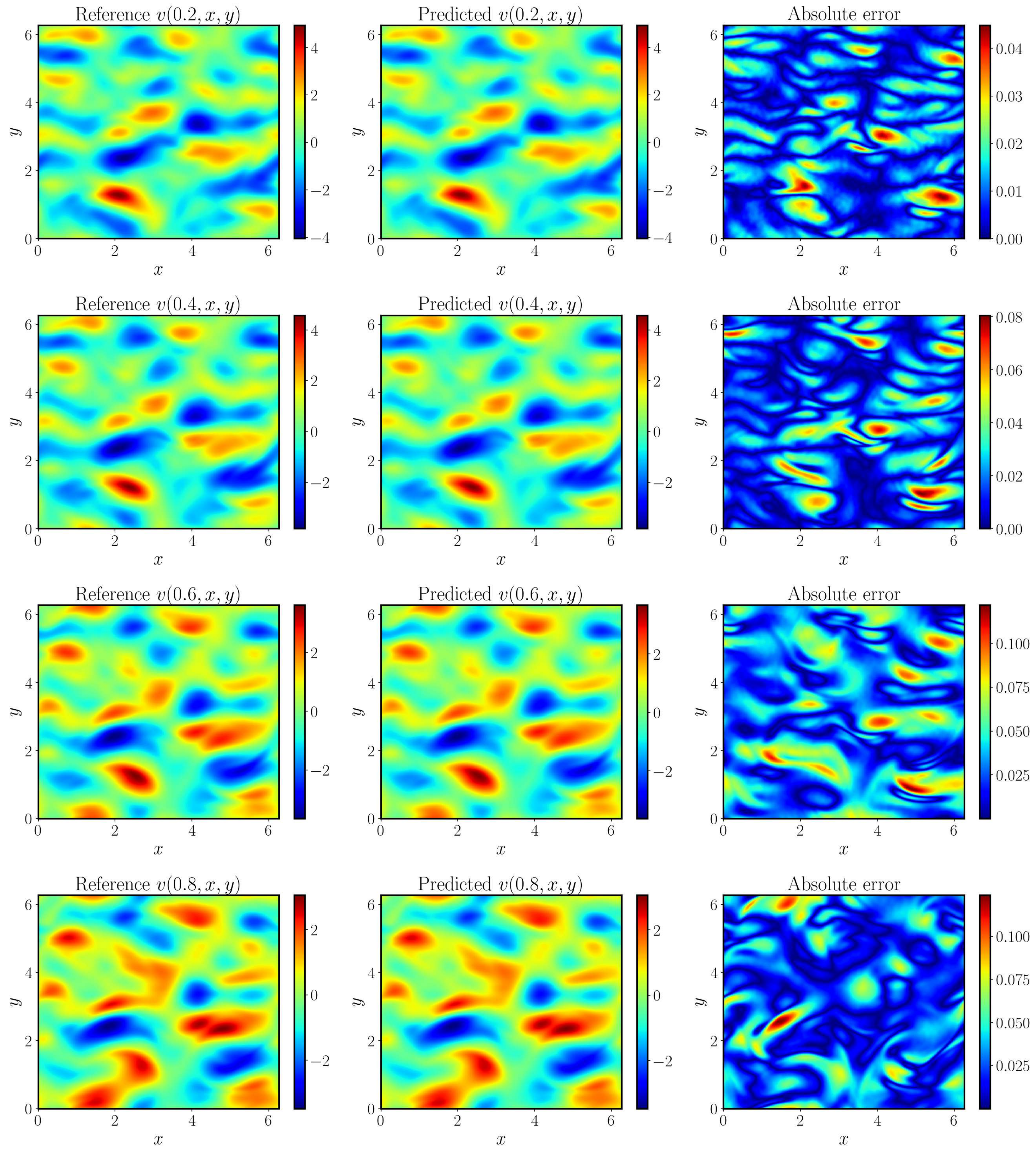}
    \caption{{\em Navier-Stokes:}  Representative snapshots of the predicted $v$ against the ground truth at $t=0.2, 0.4, 0.6, 0.8$.}
    \label{fig: turbulence_v_preds}
\end{figure}

\begin{figure}[h]
    \centering
    \includegraphics[width=0.8\textwidth]{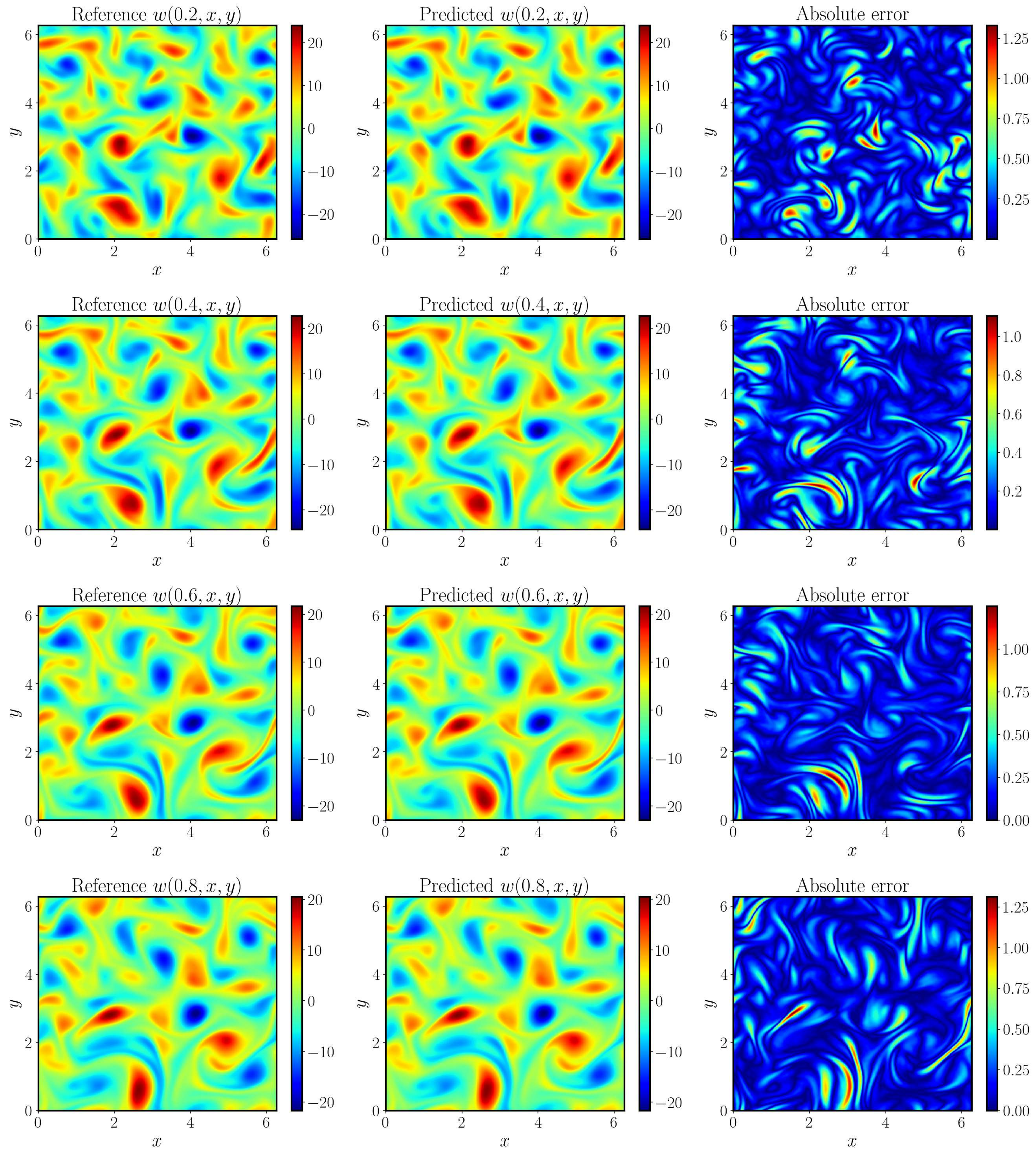}
    \caption{{\em Navier-Stokes:}  Representative snapshots of the predicted $w$ against the ground truth at $t=0.2, 0.4, 0.6, 0.8$.}
    \label{fig: turbulence_w_preds}
\end{figure}

\begin{figure}[h]
    \centering
    \includegraphics[width=0.7\textwidth]{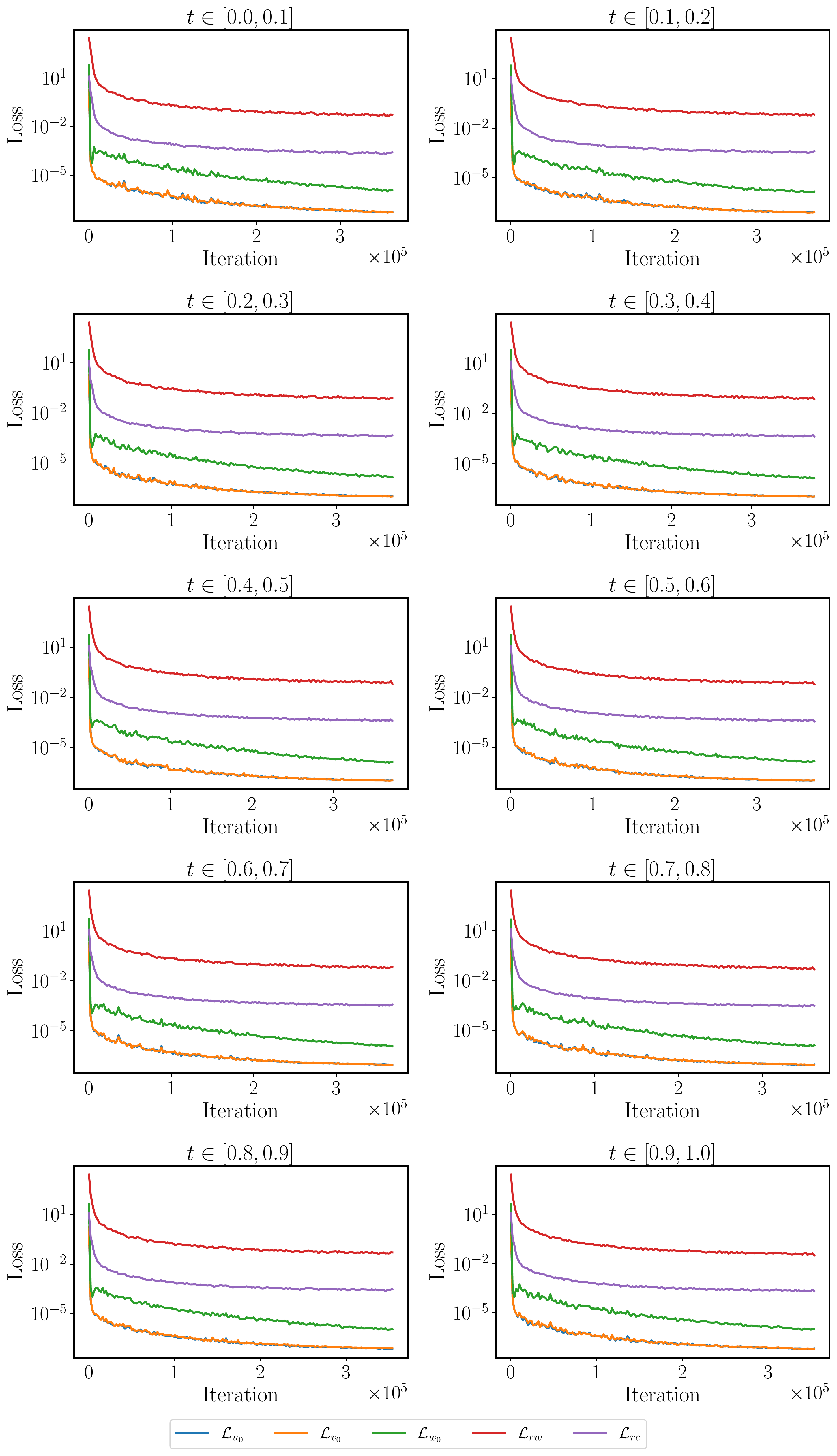}
    \caption{{\em Navier-Stokes:} Loss convergence of training a physics-informed neural network using  Algorithm \ref{alg} for every time window.}
    \label{fig: turbulence_loss}
\end{figure}

